\documentclass[lettersize,journal]{IEEEtran}

\usepackage{amsmath,amsfonts}
\usepackage{algorithmic}
\usepackage{algorithm}
\usepackage{array}
\usepackage[caption=false,font=normalsize,labelfont=sf,textfont=sf]{subfig}
\usepackage{textcomp}
\usepackage{stfloats}
\usepackage{url}
\usepackage{verbatim}
\usepackage{graphicx}
\usepackage{cite}
\usepackage{float}
\usepackage{xcolor}

\usepackage[justification=centering]{caption}

\usepackage{amsmath,amsfonts,bm}

\def\eqref#1{equation~\ref{#1}}

\def\1{\bm{1}}

\DeclareMathAlphabet{\mathsfit}{\encodingdefault}{\sfdefault}{m}{sl}
\SetMathAlphabet{\mathsfit}{bold}{\encodingdefault}{\sfdefault}{bx}{n}

\newcommand{\E}{\mathbb{E}}

\DeclareMathOperator*{\argmin}{arg\,min}

\usepackage{multirow, array, booktabs, tabularx}
\usepackage{adjustbox}
\usepackage{arydshln}
\usepackage{pifont}
\usepackage{diagbox}

\newcommand{\cmark}{\ding{51}} 
\newcommand{\xmark}{\ding{55}} 
\newcolumntype{P}[1]{>{\centering\arraybackslash}p{#1}}

\title{A Survey and Evaluation of \\ Adversarial Attacks in Object Detection}

\author{Khoi Nguyen Tiet Nguyen$^*$, Wenyu Zhang$^*$, Kangkang Lu, Yu-Huan Wu, Xingjian Zheng, \\ Hui Li Tan, Liangli Zhen
\thanks{* These two authors contributed equally to this work.}
\thanks{This work was done while K. N. T. Nguyen was with the Institute for Infocomm Research, A*STAR, Singapore. Email: nguyentietnguyenkhoi@gmail.com}
\thanks{W. Zhang, K. Lu, H. L. Tan are with the Institute for Infocomm Research, A*STAR, Singapore. Email: \{zhang\_wenyu, lu\_kangkang, hltan\}@i2r.a-star.edu.sg}
\thanks{Y.-H. Wu, X. Zheng, L. Zhen are with the Institute of High Performance Computing, A*STAR, Singapore. Email: \{wu\_yuhuan, zheng\_xingjian, zhen\_liangli\}@ihpc.a-star.edu.sg}
}

\begin{document}

\markboth{Accepted to IEEE Transactions on Neural Networks and Learning Systems (TNNLS)}%
{Shell \MakeLowercase{\textit{et al.}}: Pre-print version}

\maketitle

\begin{abstract}
Deep learning models achieve remarkable accuracy in computer vision tasks, yet remain vulnerable to adversarial examples--carefully crafted perturbations to input images that can deceive these models into making confident but incorrect predictions. This vulnerability pose significant risks in high-stakes applications such as autonomous vehicles, security surveillance, and safety-critical inspection systems. While the existing literature extensively covers adversarial attacks in image classification, comprehensive analyses of such attacks on object detection systems remain limited. This paper presents a novel taxonomic framework for categorizing adversarial attacks specific to object detection architectures, synthesizes existing robustness metrics, and provides a comprehensive empirical evaluation of state-of-the-art attack methodologies on popular object detection models, including both traditional detectors and modern detectors with vision-language pretraining. Through rigorous analysis of open-source attack implementations and their effectiveness across diverse detection architectures, we derive key insights into attack characteristics. Furthermore, we delineate critical research gaps and emerging challenges to guide future investigations in securing object detection systems against adversarial threats. Our findings establish a foundation for developing more robust detection models while highlighting the urgent need for standardized evaluation protocols in this rapidly evolving domain.
\end{abstract}


\IEEEpeerreviewmaketitle

\section{Introduction}
\label{sec: introduction}

Deep learning models have demonstrated great success in computer vision on a diverse range of tasks such as image classification, object detection, segmentation, pose estimation, and image captioning~\cite{srivastava2024omnivec, wang2023onepeace, geng2023pose}. As these models are increasingly deployed in real-world systems, it is essential to understand the model vulnerabilities that can compromise the safety and security of the systems. \textit{Adversarial examples} are carefully crafted modifications to input data that cause deep learning models to make incorrect predictions, even when these modifications are so subtle they remain imperceptible to humans~\cite{ding2020cvsurvey, huang2020cvsurvey, akhtar2018cvsurveythreat, charkraborty2021cvsurvey}. These malicious inputs pose significant safety and security concerns, particularly when deployed against high-stakes applications like autonomous vehicles, security surveillance, and safety-critical inspection systems.

\begin{table}[htb]

\caption{List of notations and acronyms used in our paper. Detailed definitions are provided in the respective sections. Refer to Table~\ref{tab: methods_literature_od} for acronyms of attack methods.
\protect\label{tab: acronyms}}

\centering
\begin{adjustbox}{max width=\columnwidth}
\begin{tabular}{ll}
\toprule[1pt]\midrule[0.3pt]

\textbf{Notations \&} & \textbf{Brief Definition}\\
\textbf{Acronyms}\\
\midrule
$x$ & Clean input image \\
\midrule
$\delta$ & Perturbation to inject to the clean image $x$\\
\midrule
$x'$ & Adversarial image defined by $x'=x+\delta$\\
\midrule
\multirow{2}{*}{$\delta^*$} & The optimal perturbation obtained after some\\
& iterative updates\\
\midrule
\multirow{2}{*}{$y$} & True label for $x$ containing $J$ objects,\\
& $y[j]=(b_x, b_y, b_h, b_w, c), j \in J$\\
\midrule
\multirow{2}{*}{$b_x,b_y, b_h, b_w$} & $b_x, b_y$ are coordinates of top-left point, $b_h, b_w$\\
&are height and width, of object bounding box $b$ \\
\midrule
$c$ & Class label of the object\\
\midrule
$\hat{y}$ & Object detector output for $x$\\
\midrule
$y'$ & Target label for $x'$\\
\midrule
\multirow{3}{*}{$\tilde{L}_{od}(x, y, \Theta)$} & Training objective of object detection model \\
& parameterized by $\Theta$, combining classification \\ 
& loss $\Tilde{L}_{clf}$, objectness loss $\Tilde{L}_{obj}$, localization \\
& regression loss $\Tilde{L}_{reg}$\\
\midrule
\multirow{2}{*}{$\epsilon$} & Perturbation budget; Threshold value for\\
& perturbation constraint such that $d(x', x) < \epsilon$ \\
& with distance function $d$\\
\midrule
$L_p$-norm & Perturbation norm measuring the amount of\\
($L_1$, $L_2$, $L_{\infty}$) & perturbation added to the image\\
\midrule
IoU & Intersection-over-Union \\
\midrule
AP@$\gamma$ & Average Precision, the precision averaged across \\
& all object classes at a fixed IoU threshold $\gamma$ \\
\midrule
mAP & Mean Average Precision, the average of AP@$\gamma$\\
& across different IoU thresholds\\
\midrule
ASR & Attack Success Rate\\
\midrule
FPR, FNR & False Positive Rate, False Negative Rate\\
\midrule
PSNR & Peak Signal-to-Noise Ratio\\
\midrule
SSIM & Structural Similarity Index Measure \\
\midrule
RPN & Region Proposal Network \\
\midrule
NMS & Non-Maximum Suppression \\
\midrule
RoI & Region of Interest potentially containing objects \\
\midrule[0.3pt]\bottomrule[1pt]
\end{tabular}
\end{adjustbox}


\end{table}

While the literature on adversarial attacks in image classification is extensive, with numerous comprehensive surveys~\cite{akhtar2018cvsurveythreat, huang2020cvsurvey, ding2020cvsurvey, charkraborty2021cvsurvey, tan2022cvsurvey, wang2022cvsurveyblackbox} and evaluations~\cite{croce2020robustbench, tang2021robustart, ren2021advret}, research examining adversarial attacks on object detection remains relatively limited. This gap is particularly significant given that object detection models present more complex attack challenges than image classification models, due to their varied network architectures, modules, and sub-processes. Existing surveys on object detection adversarial attacks have notable limitations. For instance, Amirkhani \textit{et al.} focused on object detection for the autonomous vehicle application~\cite{amirkhani2022surveyav}. Mi \textit{et al.} surveyed adversarial attacks and defences but omitted evaluation procedures and comparative attack assessments~\cite{mi2023surveyod}. 
Amongst existing evaluation studies~\cite{xu2020evaluation, hingun2022reap}, they have not addressed attack transferability in the more realistic black-box system scenarios, and some of them focused on patch attacks only while excluding other non-patch-based attacks~\cite{hingun2022reap, wu2020cloakeccv}. Furthermore, conducting a fair comparative analysis of attack method effectiveness based solely on reported results in published articles presents significant methodological challenges. These challenges stem from substantial variations across studies in multiple key dimensions: choice of detection architectures, dataset selection (full~\cite{chow2020tog, chen2022rad}, subset~\cite{cai2023ensemble, cai2022zqa}, or proprietary~\cite{xu2020evaluation}), evaluation metrics (standard mAP~\cite{chow2020tog, chen2022rad} versus custom metrics~\cite{cai2023ensemble, cai2022zqa}), attack scope (all classes~\cite{chow2020tog} versus specific targets~\cite{wang2022transpatch, wu2020cloakeccv}), and hyper-parameters like varying constraints on allowed query volumes~\cite{cai2022zqa,liang2021prfa, wang2022daedalus}.

The key factors that have contributed to the scarcity of evaluation studies on object detection-specific adversarial attacks include: 1) the limited availability of source code for many proposed methods and 2) the incompatibility of dependencies across existing open-source implementations, which prevents unified testing across different detection models. While some evaluation efforts exist, they are limited in scope. For example, Xu \textit{et al.} evaluated six adversarial attacks in terms of their effectiveness, computational cost, number of attack iterations and magnitute of image distortion~\cite{xu2020evaluation}. Similarly, Hingun \textit{et al.} evaluated how adversarial patches affect the robustness of object detectors on road signs~\cite{hingun2022reap}. Both of these studies are conducted in a white-box setting, where attackers have complete access to the detection system. In contrast, Du \textit{et al.} and Yang \textit{et al.} explored attack transferability in the black-box setting~\cite{du2022detectsec, yang2018cloak}, where attackers have limited system access. However, their scope was narrow: the first one examined only two object detection-specific attack algorithms, while another one focused on comparing different patch patterns using a single patch attack algorithm. A summary of these evaluations is reported in Table~\ref{tab: bench_object}.

\begin{table*}[htb]

\caption{Evaluation of adversarial robustness in object detection tasks. Yang et al.\ \cite{yang2018cloak} evaluated one patch attack algorithm across five patterns$^a$, while Du et al.\ \cite{du2022detectsec} examined 13 attacks, though only two specifically target object detection rather than image classification$^b$. \protect\label{tab: bench_object}}

\centering
\begin{adjustbox}{max width=0.8\textwidth}
\begin{tabular}{llllllllll}
\toprule[1pt]\midrule[0.3pt]

\textbf{Article} & \multicolumn{2}{c}{Software} & \multicolumn{2}{c}{Attack Generation} & \multicolumn{4}{c}{Robustness Evaluation}  \\
\cmidrule(lr){2-3}    
\cmidrule(lr){4-5}
\cmidrule(lr){6-9}   

& Framework & Open-source & Knowledge & Norm  & \# Detectors & \# Attacks & Metric & Datasets \\
\midrule

\multirow{ 2}{*}{ \cite{xu2020evaluation} }
& \multirow{ 2}{*}{Not specified} & \multirow{ 2}{*}{N} & \multirow{ 2}{*}{White} & \multirow{ 2}{*}{$L_2$} & 2 one-stage  & 3 & mAP, \# iterations, & \multirow{ 2}{*}{VOC, proprietary video} \\
&   &   &   &   &  1 two-stage & 3 & time cost, distortion & \\ \midrule

\multirow{ 2}{*}{ \cite{hingun2022reap} }
& \multirow{ 2}{*}{PyTorch} & \multirow{ 2}{*}{Y} & \multirow{ 2}{*}{White} & \multirow{ 2}{*}{$L_2$} & 1 one-stage & 1 & \multirow{ 2}{*}{ASR, FNR} & \multirow{ 2}{*}{Mapillary, Traffic Sign}   \\
&   &   &   &   & 1 two-stage & 1 & & \\ \midrule

\multirow{ 2}{*}{ \cite{du2022detectsec} }
& \multirow{ 2}{*}{Tensorflow} & \multirow{ 2}{*}{N} & \multirow{ 2}{*}{White/Black} & \multirow{ 2}{*}{$L_p$} & 6 one-stage  & 2+11$^b$ & \multirow{ 2}{*}{mAP} & \multirow{ 2}{*}{VOC, COCO}   \\
&   &   &   &   &  12 two-stage & 2+11$^b$ &  & \\ \midrule

\multirow{ 2}{*}{ \cite{yang2018cloak} }
& \multirow{ 2}{*}{PyTorch} & \multirow{ 2}{*}{N} & \multirow{ 2}{*}{White/Black} & \multirow{ 2}{*}{-} & 2 one-stage  & 1$^a$ & \multirow{ 2}{*}{mAP} & \multirow{ 2}{*}{VOC, COCO, Inria Person}   \\
&   &   &   &   &  2 two-stage & 1$^a$ &  & \\

\midrule[0.3pt]\bottomrule[1pt]
\end{tabular}
\end{adjustbox}

\end{table*}

In this work, we address these limitations by providing a comprehensive survey and evaluation of adversarial attacks in object detection. Our main contributions are as follows:
\begin{itemize}
    \item We propose a novel taxonomy of adversarial attacks in object detection. This taxonomy provides a structured framework for categorizing existing attack methods and characterizing their key properties, enabling researchers to better understand the relationships and distinctions between different approaches.
    \item We conduct a comprehensive analysis of evaluation methodologies in the field, examining the various metrics, benchmark datasets, and model architectures commonly used to assess attack effectiveness. This analysis highlights both standard practices and potential limitations in current evaluation approaches.
    \item We perform a systematic evaluation of open-source attack methods and adversarial robustness of popular object detection models, including both traditional detectors and modern detectors with vision-language pretraining. Our systematic evaluation provides unprecedented insights into the relative strengths of different attack strategies and the comparative robustness of detection architectures under adversarial conditions.
    \item Based on our analysis and experimental results, we identify critical gaps in current research and outline promising directions for future work. These include developing more effective adversarial attacks for modern object detectors, designing robust defense mechanisms specifically for small object detection, preventing multimodal adversarial attacks, and advancing the state of physical adversarial attacks that can reliably fool detectors in real-world scenarios.
\end{itemize}
\section{Taxonomy of adversarial attacks}
\label{sec: taxonomy_od}

\begin{figure*}[htb]
    \centering
    \captionsetup{justification=centering}
    \includegraphics[width=0.95\textwidth]{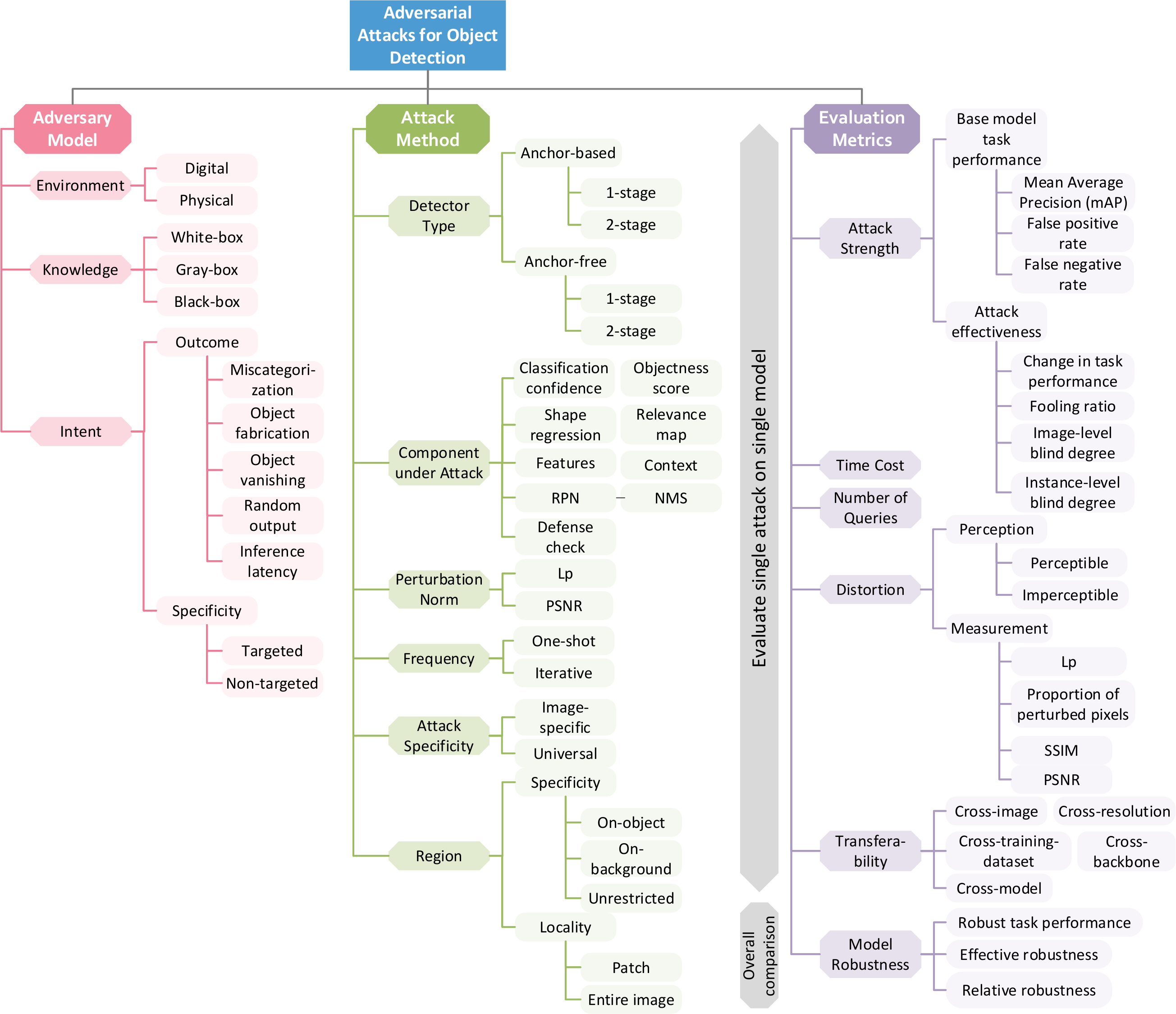}
    \caption{Taxonomy and evaluation metrics on adversarial attacks in object detection.}
    \label{fig:taxonomy_od}
\end{figure*}

We present our taxonomy of adversarial attacks in object detection in Figure~\ref{fig:taxonomy_od}. We describe preliminary information in Section~\ref{subsec: preliminaries}, followed by detailed categorizations for adversarial model and attack method in Section~\ref{subsec: adversarial model} and \ref{subsec: attack model}, respectively. We note that some taxonomy categories, namely environment, knowledge of adversary, intent specificity, perturbation norm, attack frequency and attack specificity, are common concepts that are also used in other tasks including image classification. We discuss these taxonomy categories and provide the relevant references specifically in the context of object detection. For better readability, we provide a list of notations and acronyms used in our paper in Table \ref{tab: acronyms}.

\subsection{Preliminaries}
\label{subsec: preliminaries}

The aim of object detection is to detect, localize and classify objects of interest in an image. Object detectors with real-time inference speed can also be used for streaming video. Denote $(\mathcal{X},\mathcal{Y})$ as the input-output space, and $x\in\mathcal{X}$ is an input image, and $y\in\mathcal{Y}$ is the output. For an image containing $J$ objects, $y\in\mathbb{R}^{J\times 5}$ with $y[j]=(b_x,b_y,b_h,b_w,c)$ where $(b_x,b_y)$ are the top-left corner coordinates and $(b_h,b_w)$ are the height and width of the object bounding box $b$, and $c$ is the label out of $C$ object classes. For an object detection model $f$ parameterized by $\Theta$, its prediction for $x$ is $\hat{y}=f(x;\Theta)$. Some models output $(b_x,b_y)$ as the center coordinates, and output additional information such as the confidence score $s=(s_1,\dots,s_C)$ where $s_c$ is the predictive probability for object class $c$, and objectness score $p$ of whether an object exists in the predicted box.

The training objective $\tilde{L}_{od}(x, y, \Theta)$ for object detection is typically the weighted combination of the following components for a predicted box \cite{arani2022odsurvey, Chow2020UnderstandingOD}:
\begin{itemize}
    \item Classification loss $\Tilde{L}_{clf}(x,y;\Theta) = -\sum_{i=1}^C s_i \log(\hat{s}_i)$. Besides binary cross entropy, other choices include multi-class cross entropy and focal loss.
    \item Objectness loss $\Tilde{L}_{obj}(x,y;\Theta) = - p\log(\hat{p}) - (1 - p)\log(1-\hat{p})$.
    \item Localization regression loss $\Tilde{L}_{reg}(x,y;\Theta) = (b_x - \hat{b}_x)^2 + (b_y - \hat{b}_y)^2 + (b_h - \hat{b}_h)^2 + (b_w - \hat{b}_w)^2$. Besides squared error, other choices include L1 and smooth L1 losses.
\end{itemize}

An adversarial attack is the addition of perturbation $\delta$ to image $x$ such that model prediction on the perturbed image $x'=x+\delta$ is incorrect \textit{i.e.}, $f(x';\Theta) = f(x+\delta;\Theta)\neq y$. Denoting the adversarial loss as $L_{adv}(x+\delta, y ;\Theta)$ with $f$ as the target model, the attacker search for the optimal perturbation $\delta^*$ by
\begin{equation}
    \delta^* = \argmin_\delta L_{adv}(x+\delta, y ;\Theta).
\end{equation}
For controlling the amount of perturbation added, a perturbation constraint can be added \textit{i.e.}, $d(x+\delta,x)<\epsilon$ with distance function $d(a,b)$ and threshold $\epsilon>0$.

\subsection{Adversarial model}
\label{subsec: adversarial model}

\subsubsection{Environment}

\noindent \textbf{Digital:} Digital attacks alter the pixels of images input into the target model. The attacker has no access to the physical environment and the image capture and pre-processing procedures. We focus on digital attacks in this literature review, as physical attacks are often extensions of digital attacks by additionally considering variations in physical conditions, \textit{e.g.}, lighting, object pose, camera angle.

\noindent \textbf{Physical:} 
While we focus on digital attacks in the literature review, we note that some papers implemented physical versions of their proposed attacks as sticker attacks \cite{yang2018cloak}. A common strategy is to use Expectation Over Transformation (EOT) \cite{Athalye2017eot, yang2018cloak} which augments the input image $x$ with a sampled set of transformations $\{t\}$ from distribution $T$ to simulate real-world transformations such as lighting and viewing angle changes, and optimizes the adversarial image $x'=x+\delta$ over the expectation of the transformed images with the EOT loss:
\begin{equation}
    \delta^* = \argmin_\delta \E_{t\sim T} L_{adv}(x+\delta, y;\Theta)
\end{equation}
subject to $\E_{t\sim T} d(t(x+\delta), t(x)) < \epsilon$ with distance function $d(a, b)$ and threshold $\epsilon$.

Thys \textit{et al.} proposed to minimize a non-printability score to favor pixel values close to a set of printable colors $Color_{print}$~\cite{thys2019surveillance}:
\begin{equation}
    L_{nps}(x') = \sum_{i,j} \min_{color\in Color_{print}} | x'_{i,j} - color|
\end{equation}
where $x'_{i,j}$ is the pixel at location $(i,j)$. Non-printability refers to the challenge of generating adversarial attacks that remain effective after printing and affixing on real-world objects, by considering physical constraints such as color gamut limitations, resolution and detail loss, and lighting and other environmental conditions.

Thys \textit{et al.} also considered to minimize a total variation loss to produce images with smooth color transitions instead of noisy pixels which may not be captured well by detectors in physical environments~\cite{thys2019surveillance} :
\begin{equation}
    L_{tv}(x') = \sum_{i,j} \sqrt{(x'_{i,j} - x'_{i+1,j})^2 + (x'_{i,j} - x'_{i,j+1})^2}.
\end{equation}

\begin{figure*}[htb]
    \centering
    \captionsetup{justification=centering}
    \includegraphics[width=0.8\linewidth]{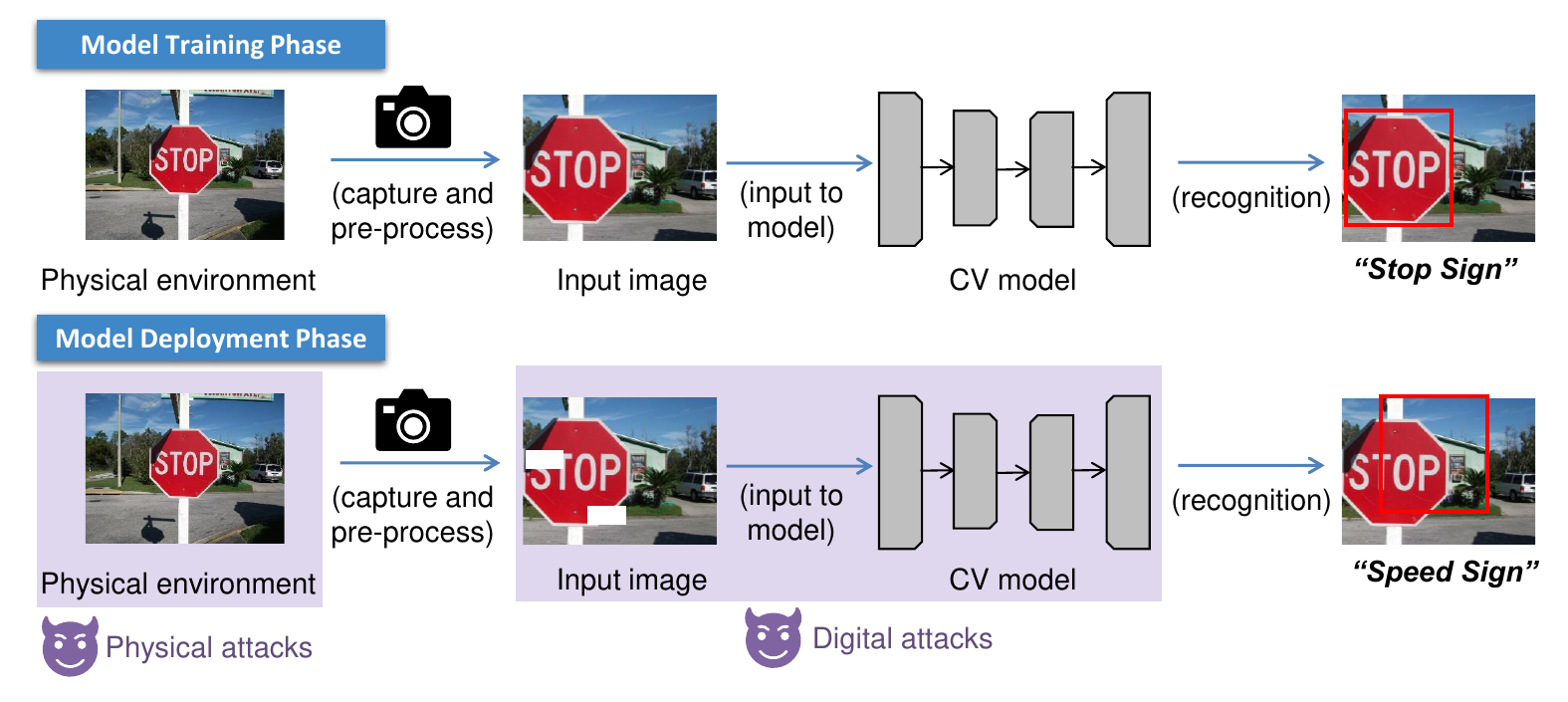}
    \caption{Adversarial attack procedure.}
    \label{fig:environment}
\end{figure*}

The evaluation of attacks and defenses in the real world is exceptionally costly. Hingun \textit{et al.} utilized more realistic simulation to study the effect of real-world patch attacks on road signs, by simulating physical adversarial stickers subject to different locations, orientations and lighting conditions~\cite{hingun2022reap}. However, they did not evaluate the fidelity of the simulated images to real-world images.

\subsubsection{Knowledge}

\noindent \textbf{White-box attacks:} 
In a white-box attack, the adversary has complete knowledge of the target model. Information includes the model input and output, neural network architecture, parameter weights and gradients, decision-making process, and training procedure and dataset. The attacker either has full access to the target model or has the capability to completely reconstruct the target model. With this level of information, the adversary can craft attacks specific to the target model.

\noindent \textbf{Gray-box attacks:} 
In a gray-box attack, the adversary has some but incomplete knowledge of the target model. For instance, the adversary may not have knowledge of the network architecture or training dataset \cite{xie2017dag}. For defended models, Yin \textit{et al.} assumed that the adversary knows the defense neural network architecture but not the parameters~\cite{yin2022adc}. In the gray-box setting, the adversary can use a surrogate model similar to the target model to craft attacks instead \cite{xie2017dag, yin2022adc}. Due to the limitation in information available, gray-box attacks are typically less effective than white-box attacks.

\noindent \textbf{Black-box attacks:} 
In a black-box attack, the adversary has no knowledge of the target model. The adversary may submit input queries and observe the predicted bounding boxes and class confidence scores \cite{liang2021prfa}. Common models and algorithms can be used as surrogates as they may share vulnerabilities with the target model \cite{wang2022daedalus}. Cai \textit{et al.}  utilized an ensemble of $M$ surrogate models and proposed to minimize a joint adversarial loss~\cite{cai2022context}:
\begin{equation}
    \begin{aligned}[b]
    & L_{adv-ensemble}(x+\delta, y; \{\alpha_m\}, \{\Theta_m\}) =\\
    & \sum_{m=1}^M \alpha_m L_{adv}(x+\delta, y ;\Theta_m)
    \end{aligned}
\end{equation}
where $L_{adv}(x+\delta, y ;\Theta_m)$ is the loss for the $m$-th surrogate model $f_m$ parameterized by $\Theta_m$, $\alpha_m > 0$ for $m\in\{1,\dots,M\}$ and $\sum_{m=1}^M \alpha_m = 1$.
Black-box attacks are the most challenging attacks to execute effectively. However, due to the lack of assumption on target model information, they are versatile to implement and are applicable to a wide range of models.

\subsubsection{Intent outcome}

Integrity-based attacks aim to compromise the accuracy of the model predictions \cite{Chow2020UnderstandingOD}.

\noindent \textbf{Random output:} The adversary perturbs the input such that the model detection output is different from the ground-truth, but does not have a specific intended outcome on the accuracy of the predicted bounding box or object label. An example is to learn $\delta$ by maximizing the detection training objective $\tilde{L}_{od}(x+\delta,y;\Theta)$.

\noindent \textbf{Object vanishing:} The goal of the adversary is for the model to miss detecting objects in the input image. For examples, \cite{yang2018cloak} and \cite{thys2019surveillance} attack such that no human is detected in the inputs. In \cite{chen2022rad}, bounding boxes can be shrunk in size. The adversary can learn $\delta$ to reduce objectness scores of candidate boxes.

\noindent \textbf{Object fabrication:} The goal of the adversary is for the model to produce redundant bounding boxes or to detect non-existent objects in the input image. The adversary can learn $\delta$ to increase objectness scores of candidate boxes.

\noindent \textbf{Miscategorization:} The goal of the adversary is to perturb the input such that the model predicts an incorrect label for a detected object. Using the training objective for object detection as the adversarial loss, the adversary can learn $\delta$ by minimizing $\tilde{L}_{od}(x+\delta,y';\Theta)$ with target class label $y' \neq c$.

\begin{figure*}[htb]
    \centering
    \includegraphics[width=0.9\linewidth]{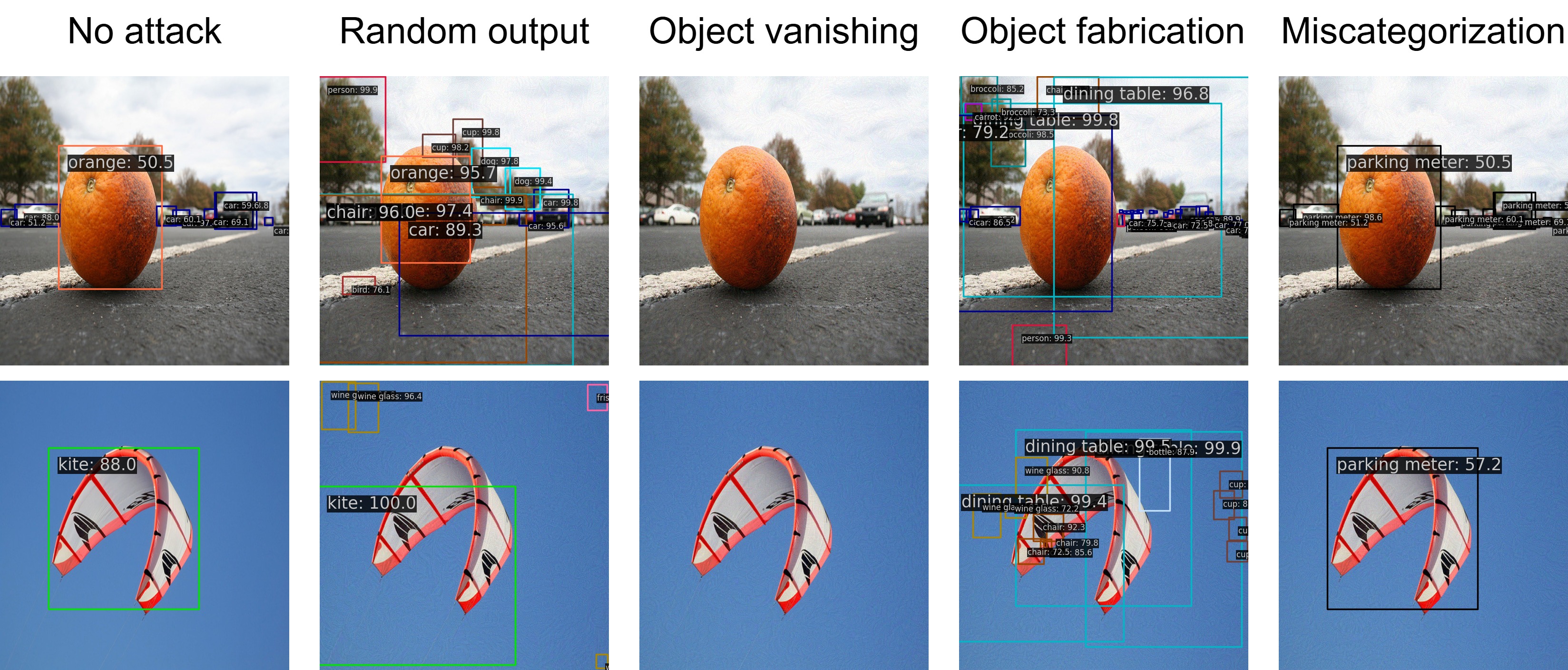}
    \caption{Examples of outcomes of integrity-based attacks. Original images taken from MS COCO 2017 \cite{Lin2014MicrosoftCC}}
    \label{fig:examples}
\end{figure*}

Availability-based attacks aim to compromise uninterrupted access to the model to process inputs.

\noindent \textbf{Inference latency}: The goal of the adversary is to increase the model's inference time, which can disrupt real-time applications. \cite{shapira2023phantom} fabricate boxes for non-existent objects to overload the Non-Maximum Suppression (NMS) \cite{redmon2016yolo} algorithm.

\subsubsection{Intent specificity}

\noindent \textbf{Targeted:} In targeted attacks, the adversary perturbs the input such that for the attacked object, the model assigns it the class label specified by the adversary. In object vanishing attacks, the target is the background class or the non-detection of objects existent in the image. 

\noindent \textbf{Non-targeted:} In non-targeted attacks, the adversary do not specify class labels to be predicted for attacked objects.

\subsection{Attack method}
\label{subsec: attack model}

\subsubsection{Detector type}

We categorize object detectors by their usage of anchor boxes and the number of stages in the algorithm.

Anchor-based techniques are commonly used in deep learning object detection algorithms. Anchor-based detectors create anchor boxes of various shapes and sizes placed at different locations on the image to represent the objects to be detected. The anchor boxes act as priors and are refined as the object detection network learns to predict the objectness of the anchor boxes, offset of the anchor boxes from the ground-truth boxes, as well as the object class labels. Redundant boxes are removed to output the final predicted bounding boxes.

\noindent \textbf{One-stage anchor-based:} One-stage anchor-based detectors directly predicts the bounding box coordinates and class labels of objects in an image. Examples include SSD \cite{Liu2015SSD} which use pre-defined anchor boxes, and YOLO v4 \cite{Bochkovskiy2020YOLOv4} which learns anchor boxes during training.

\noindent \textbf{Two-stage anchor-based:} In the first stage, a set of candidate object regions is generated by a region proposal algorithm such as a region proposal network (RPN) that utilizes anchor boxes \cite{Ren2015FasterRCNN}. In the second stage, a neural network refines and classifies the region proposals. Examples include Faster R-CNN \cite{Ren2015FasterRCNN} and Libra R-CNN \cite{pang2019libraRCNN}.

\cite{Wu2020DPAttackDP} empirically finds that it is more difficult to attack two-stage detectors as compared to one-stage detectors. Two-stage detectors like Faster R-CNN have features with a smaller receptive field than one-stage detectors like YOLO v4, which contribute to increased robustness to local perturbations.

A disadvantage of anchor-based detectors is the need to process a large number of anchor boxes in order to cover objects of different shapes and sizes, which can increase computation time. Anchor-free methods do not need anchor boxes as priors, and hence can do away with hand-crafted components for anchor generation.

\noindent \textbf{One-stage anchor-free:} One-stage anchor-free detectors directly predict the coordinates of the bounding boxes. Examples include YOLO \cite{redmon2016yolo}, FoveaBox \cite{kong2019foveabox} and FCOS \cite{tian2022fcos}. Apart from CNN-based models, transformers-based models such as DETR \cite{carion2020detr} and Deformable DETR \cite{zhu2020deformable} are more recently developed.
They further simplified the detection process, directly predicting tokens as bounding boxes, whose process is supervised via Hungarian matching. Therefore, they do not rely on non-maximum suppression that most CNN-based detectors applied on.

\noindent \textbf{Two-stage anchor-free:} Similar to two-stage anchor-based methods, a region proposal algorithm generates candidate object regions in the first stage, and the region proposals and refined in the second stage. The region proposal algorithm, such as Selective Search \cite{girshick2013rcnn}, does not make use of anchor boxes. Examples of such detectors include R-CNN \cite{girshick2013rcnn},  Corner Proposal Network \cite{Duan2020CornerPN}, and Sparse R-CNN \cite{sun2021sparse}.

\subsubsection{Component under attack}

An adversarial method can attack a combination of components to achieve the desired outcome of the adversary. 

\noindent \textbf{Classification confidence score}:
From our summary Table~\ref{tab: methods_component_under_attack}, most methods attack the classification confidence score $s$. Attacks achieve misclassification in detection by increasing the predictive probability of an incorrect label over that of the correct label \cite{xie2017dag, chow2020tog, Wu2020DPAttackDP}. UEA \cite{wei2019uea} uses a Generative Adversarial Network (GAN) to generate perturbations that produce inaccurate confidence scores. 

For models that do not estimate a separate objectness score, object class confidence is used as objectness score. For object vanishing attacks, object class confidence is lowered \cite{yang2018cloak, hao2021rpattack, liang2021prfa} and background score is raised \cite{li2021udos, Li2018ExploringTV, wu2019guap}. For object fabrication attacks, object class confidence is increased \cite{shapira2023phantom}.

\noindent \textbf{Objectness score}:
The objectness scores of predicted boxes can be increased for object fabrication \cite{chow2020tog} or decreased for object vanishing \cite{chow2020tog, wang2022transpatch, thys2019surveillance} attacks. TransPatch \cite{wang2022transpatch} utilizes a transformer generator to create perturbations.

\noindent \textbf{Shape regression}:
By attacking the shape regression module, predicted box location and size becomes inaccurate \cite{Liu2019DPATCH, chow2020tog, Li2018ExploringTV, li2018rap}. PRFA \cite{liang2021prfa} reduce the Intersection-over-Union (IoU) of predicted and ground-truth boxes. Daedalus \cite{wang2022daedalus} and \cite{shapira2023phantom} compress dimensions of predicted boxes to create more redundant boxes.

\noindent \textbf{Relevance map}:
Authors for RAD \cite{chen2022rad} observes that relevance maps from detection interpreters is common across detectors and such that a wider range of detectors are vulnerable to relevance map attacks in black-box setting.

\noindent \textbf{Features}:
UEA \cite{wei2019uea} regularizes features of foreground objects to be random values to damage object information in the features.

\noindent \textbf{Context}:
CAP \cite{zhang2020CAP} damages contextual information in an identified Region of Interest (RoI) by decreasing object classification scores and increasing background scores of the contextual region. Based on the co-occurrence object relation graph of the victim object, \cite{cai2022context} identifies helper objects that typically co-occur with the victim object, add perturbations to modify labels of both the victim and helper objects to increase attack success rate.

\noindent \textbf{Region proposal algorithm}:
In two-stage detectors, when quality of proposals from the region proposal algorithm in the first stage is degraded, consequent detection performance in the second stage will be affected.
G-UAP \cite{wu2019guap} decreases confidence score of foreground and increases confidence score of background so that the region proposal network (RPN) mistakes foreground for background. R-AP \cite{li2018rap} degrades RPN performance by reducing objectness score and disturbing shape regression.

\begin{figure}[htb]
    \centering
    \includegraphics[width=\columnwidth]{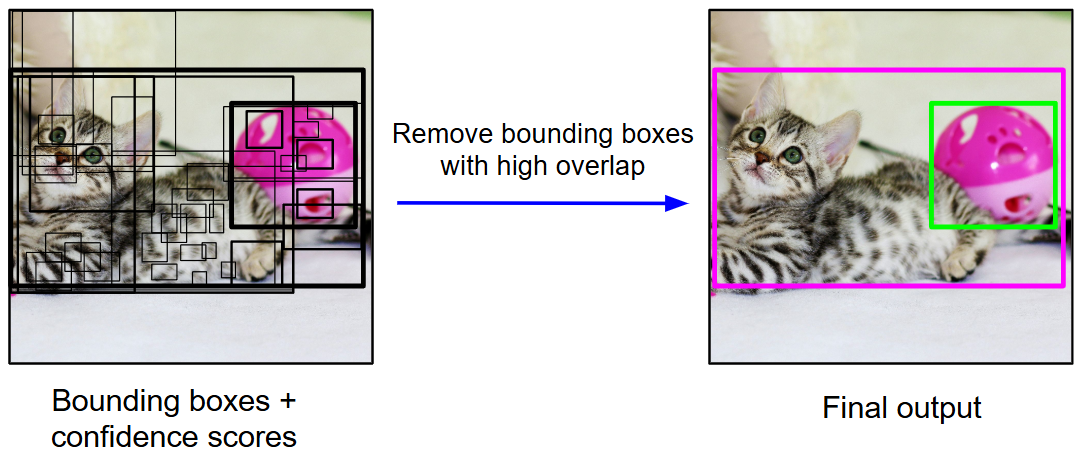}
    \caption{Non-Maximum Suppression (NMS) a post-processing technique to remove redundant bounding boxes generated by the object detector. Original image taken from MS COCO 2017 \cite{Lin2014MicrosoftCC}.}
    \label{fig:nms}
\end{figure}

\noindent \textbf{Non-Maximum Suppression}:
Non-Maximum Suppression (NMS) \cite{redmon2016yolo} is a post-processing technique to remove redundant bounding boxes generated by the object detector. For a given object class, the detected bounding box with the highest confidence score above a specified threshold is selected, and other bounding boxes with overlap above a specified IoU threshold are discarded. Figure~\ref{fig:nms} shows the bounding boxes before and after NMS post-processing in YOLO \cite{redmon2016yolo}.

Daedalus \cite{wang2022daedalus} fabricate non-existent objects by reducing the number of bounding boxes filtered off by NMS. Perturbations are added to maximize confidence scores of the detected bounding boxes, and to minimize IoU for each pair of boxes. \cite{shapira2023phantom} seeks to increase inference time by increasing detected bounding boxes to load the NMS module.

\noindent \textbf{Defence check}:
Context consistency checks aim to detect attacks through inconsistent object-to-object co-occurrence relationships that can arise in an attacked image. ADC \cite{yin2022adc} and ZQA \cite{cai2022zqa} can evade context consistency checks by assigning class labels consistent with normal co-occurrence relationships.

\subsubsection{Perturbation norm}

A perturbation constraint can be added to the adversarial loss to control the amount of perturbation added, such that the perturbations are visually imperceptible to human observers. Typically, the $L_p$-norm, $\|\delta\|_p$ is used, with $p\in\{0,1,2,\infty\}$. CAP \cite{zhang2020CAP} uses the Peak Signal-to-Noise Ratio (PSNR):
\begin{equation}
    PSNR = 10 \log_{10} \frac{MAX(x)^2}{MSE(x, x+\delta)} \label{eqn: psnr}
\end{equation}
where $MAX(x)$ is the maximum pixel value in $x$, and $MSE(x, x+\delta)=\frac{1}{h(x)w(x)} \|\delta\|_2^2$ for image $x$ with height $h(x)$ and width $w(x)$.

\subsubsection{Frequency}

\noindent \textbf{One-shot:}
In a one-shot attack, the adversary has a single attempt to perturb the input image. One-shot attacks may be less effective than iterative attacks, but are typically more time and resource-efficient, and consequently more suitable for real-time attacks. Such perturbations can be generated by Generative Adversarial Networks (GANs) \cite{wei2019uea}. ZQA \cite{cai2022zqa} estimates a perturbation success probability matrix to select the list of victim objects and labels that is most likely to lead to a successful attack, and does not iteratively query the target model to check the outcome of the attack.

\noindent \textbf{Iterative:}
In an iterative attack, the adversary can repeatedly optimize the perturbation based on feedback from the target model. Many existing methods use iterative gradient updates, such as projected gradient descent (PGD) \cite{madry2018resistant}, to generate a successful attack based on target model gradients. Black-box methods query the target model to check attack success and to modify the attack plan corresponding to the model output \cite{cai2022context, cai2023ensemble}. Iterative attacks typically use more time and resources than one-shot attacks.

\subsubsection{Attack specificity}

\noindent \textbf{Image-specific:}
A separate perturbation needs to be learned for each input image for image-specific attacks. These attacks are generally more effective than universal attacks.

\noindent \textbf{Universal:}
An universal attack is a perturbation learned to be applicable for any input image. While more challenging to construct than image-specific attacks, effective universal attacks pose a more significant threat as they are transferable across images without further accessing the target model. Universal perturbations are learned using a training set of images typically similarly distributed to the test images \cite{li2021udos}. Additional image-specific finetuning on universal perturbations can be applied to improve their effectiveness \cite{wu2019guap}.

\subsubsection{Region specificity}

\noindent \textbf{On-object:}
Some attacks focus their perturbations on pixels of foreground objects to cause miscategorization and object vanishing. The objects can be manually located \cite{yang2018cloak}, or objectness of pixels can be estimated through detector features or outputs \cite{wei2019uea, liang2021prfa, wang2022transpatch}. On-object perturbations can be constructed into stickers, posters, clothes, or other surfaces to be affixed to the object for physical attacks \cite{yang2018cloak}.

\noindent \textbf{On-background:} Some methods focus their perturbations on the background. For example, Li \textit{et al.} increased the number of false positives by perturbing the background~\cite{Li2018ExploringTV}.

\noindent \textbf{Unrestricted:} Without placing a restriction on the type of pixels under attack, the entire image can be perturbed.

\subsubsection{Region locality}

\noindent \textbf{Patch:} Patch attacks constrain perturbations within region of a fixed shape, typically a rectangle. Some methods allow multiple patches \cite{liang2021prfa, hao2021rpattack}. 
Patches of pre-specified sizes can be constructed into stickers for physical attacks \cite{yang2018cloak, wang2022transpatch, thys2019surveillance}.

\noindent \textbf{Entire image:} Without constraining perturbations to pixels within a patch, the entire image can be perturbed. A perturbation norm on $\delta$ is typically used to constraint the magnitude of perturbation.

We summarize a list of articles on adversarial attack methods in object detection in Table~\ref{tab: methods_literature_od} and \ref{tab: methods_component_under_attack}. We base the knowledge of the adversary model on the method description as well as the transferability experiments, that is, we include gray and/or black-box attack capabilities for a white-box attack if the article demonstrates that the attack has high transferability despite limited knowledge of the target model. We include gray-box capability for cross-resolution, cross-training-dataset or cross-backbone transferability, and black-box capability for cross-model transferability. Description on transferability types can be found in Section~\ref{subsec: transferability}.

\begin{table*}[!ht]

\caption{Articles on adversarial attack methods in object detection. $^*$ denotes open-source code is available. For detector type, 1-AB, 2-AB, 1-AF, 2-AF denote one-stage anchor-based, two-stage anchor-based, one-stage anchor-free and two-stage anchor-free detectors, respectively. \protect\label{tab: methods_literature_od}}

\centering
\begin{adjustbox}{max width=\textwidth}
\begin{tabular}{l*{2}{c}P{3cm}P{2cm}P{2cm}P{2cm}P{2.2cm}*{3}{c}}
\toprule[1pt]\midrule[0.3pt]

\textbf{Article} & \multicolumn{4}{c}{\textbf{
Adversary Model}} & \multicolumn{6}{c}{\textbf{
Attack Model}} \\ \cmidrule(lr){2-5} \cmidrule(lr){6-11}
& \textbf{Environment} & \textbf{Knowledge} & \textbf{Outcome} & \textbf{Intent Specificify} & \textbf{Detector Type} & \textbf{Frequency} & \textbf{Attack Specificity} & \textbf{Perturbation Constraint} & \textbf{Region Specificity} & \textbf{Locality} \\ \midrule
DAG$^*$ \cite{xie2017dag} & Digital & White/Gray/Black & Miscategorization & Targeted & 2-AB & Iterative & Image-specific & - & Unrestricted & Entire image \\
Li et al. \cite{Li2018ExploringTV} & Digital & White/Gray & Random output & Non-targeted & 1-AB/2-AB & Iterative & Image-specific & - & On-background & Patch \\
Yang et al. \cite{yang2018cloak} & Digital/Physical & White & Object vanishing & Targeted & 1-AB & Iterative & Image-specific & - & On-object & Patch \\
R-AP$^*$ \cite{li2018rap} & Digital & Black & Random output & Non-targeted & 2-AB & Iterative & Image-specific & Peak Signal-to-Noise Ratio & Unresticted & Entire image \\
UEA$^*$ \cite{wei2019uea} & Digital & White & Random output & Non-targeted & 1-AB/2-AB & One-shot & Image-specific & - & On-object & Entire image \\
DPatch$^*$ \cite{Liu2019DPATCH} & Digital & White/Gray/Black & Random output & Targeted/ Non-targeted & 1-AB/2-AB & Iterative & Image-specific & - & Unrestricted & Patch \\
G-UAP \cite{wu2019guap} & Digital & White/Gray/Black & Object vanishing & Targeted & 2-AB & Iterative & Universal & $L_\infty$ & Unrestricted & Entire image \\
Thys et al. \cite{thys2019surveillance} & Digital/Physical & White & Object vanishing & Targeted & 1-AB & Iterative & Image-specific & - & On-object & Patch \\
CAP \cite{zhang2020CAP} & Digital & White & Object vanishing & Targeted & 2-AB & Iterative & Image-specific & Peak Signal-to-Noise Ratio & Unrestricted & Entire image \\
TOG$^*$ \cite{chow2020tog} & Digital & White/Gray/Black & Miscategorization/ Object fabrication/ Object vanishing & Targeted & 1-AB & Iterative & Image-specific/ Universal & $L_0$/$L_2$/$L_\infty$ & Unrestricted & Entire image \\
DPAttack$^*$ \cite{Wu2020DPAttackDP} & Digital & White & Object vanishing & Targeted & 1-AB/2-AB & Iterative & Image-specific & - & On-object & Patch \\
Evaporate Attack \cite{wang2020evaporate} & Digital & Black & Object vanishing & Targeted & 1-AB/2-AB & Iterative & Image-specific & $L_2$ & Unrestricted & Entire image \\
U-DOS \cite{li2021udos} & Digital & White/Gray/Black & Object vanishing & Targeted & 1-AB/2-AB & Iterative & Universal & $L_\infty$ & Unrestricted & Entire image \\
RPAttack$^*$ \cite{hao2021rpattack} & Digital & White & Object vanishing & Targeted & 1-AB/2-AB & Iterative & Image-specific & - & Unrestricted & Patch\\
PRFA$^*$ \cite{liang2021prfa} & Digital & Black & Random output & Non-targeted & 1-AB/2-AB/ 1-AF & Iterative & Image-specific & $L_p$ & On-object & Patch \\
RAD$^*$ \cite{chen2022rad} & Digital & Black & Random output & Non-targeted & 1-AB/2-AB & One-shot & Image-specific & $L_\infty$ & Unrestricted & Entire image\\
ADC \cite{yin2022adc} & Digital & White/Gray & Miscategorization/ Object fabrication/ Object vanishing & Targeted & 2-AB & Iterative & Image-specific & $L_\infty$ & Unrestricted & Entire image \\
Daedalus$^*$ \cite{wang2022daedalus} & Digital & White & Object fabrication & Non-targeted & 1-AB & Iterative & Image-specific & $L_0$/$L_2$ & Unrestricted & Entire image \\
CAT $^*$ \cite{cai2022context} & Digital & Black & Miscategorization & Targeted & 1-AB/2-AB/ 1-AF & Iterative & Image-specific & $L_\infty$ & Unrestricted & Entire image\\
TransPatch \cite{wang2022transpatch} & Digital/Physical & White & Object vanishing & Targeted & 1-AB/ 1-AF & Iterative & Image-specific & - & On-object & Patch\\
ZQA \cite{cai2022zqa} & Digital & Black & Miscategorization & Targeted & 1-AB/2-AB/ 1-AF & One-shot & Image-specific & $l_\infty$ & Unrestricted & Entire image \\
EBAD $^*$ \cite{cai2023ensemble} & Digital & Black & Miscategorization & Targeted & 1-AB/2-AB/ 1-AF & Iterative & Image-specific & $l_\infty$ & Unrestricted & Entire image \\
Shapira et al. \cite{shapira2023phantom} & Digital & White & Inference latency & Targeted & 1-AB & Iterative & Universal & $L_2$ & Unrestricted & Entire image \\
\midrule[0.3pt]\bottomrule[1pt]
\end{tabular}
\end{adjustbox}

\end{table*}
\begin{table*}[tb]

\caption{Component attacked by adversarial methods for object detection. $^*$ denotes open-source code is available. \protect\label{tab: methods_component_under_attack}}

\centering
\begin{adjustbox}{max width=\textwidth}
\begin{tabular}{l*{10}{P{2cm}}}
\toprule[1pt]\midrule[0.3pt]

\textbf{Article} & \multicolumn{8}{c}{\textbf{
Component under Attack}} \\ \cmidrule(lr){2-10}
& \textbf{Confidence Score} & \textbf{Objectness Score} & \textbf{Shape regression} & \textbf{Relevance Map} & \textbf{Features} & \textbf{Context} & \textbf{Region Proposal} & \textbf{NMS} & \textbf{Defence Check}\\ \midrule
DAG$^*$ \cite{xie2017dag} & \cmark &  &  &  &  &  &  &  & \\
Li et al. \cite{Li2018ExploringTV} & \cmark &  & \cmark &  &  &  &  &  & \\
Yang et al. \cite{yang2018cloak} & \cmark &  &  &  &  &  &  &  & \\
R-AP$^*$ \cite{li2018rap} & & \cmark & \cmark & & & & \cmark & & \\
UEA$^*$ \cite{wei2019uea} & \cmark &  &  &  & \cmark &  &  &  & \\
DPatch$^*$ \cite{Liu2019DPATCH} & \cmark &  & \cmark &  &  &  &  &  &  \\
G-UAP \cite{wu2019guap} & \cmark &  &  &  &  &  & \cmark &  & \\
Thys et al. \cite{thys2019surveillance} & \cmark & \cmark \\
CAP \cite{zhang2020CAP} & \cmark &  &  &  &  & \cmark & \cmark &  & \\
TOG$^*$ \cite{chow2020tog} & \cmark & \cmark & \cmark &  &  &  &  &  & \\
DPAttack$^*$ \cite{Wu2020DPAttackDP} & \cmark &  &  &  &  &  &  &  & \\
Evaporate Attack \cite{wang2020evaporate} & \cmark \\
U-DOS \cite{li2021udos} & \cmark &  &  &  &  &  &  &  & \\
RPAttack$^*$ \cite{hao2021rpattack} & \cmark &  &  &  &  &  &  &  & \\
PRFA$^*$ \cite{liang2021prfa} & \cmark &  & \cmark &  &  &  &  &  & \\
RAD$^*$ \cite{chen2022rad} &  &  &  & \cmark &  &  &  &  & \\
ADC \cite{yin2022adc} & \cmark &  &  &  & \cmark &  &  &  & \cmark \\
Daedalus$^*$ \cite{wang2022daedalus} & \cmark &  & \cmark &  &  &  &  & \cmark & \\
CAT $^*$ \cite{cai2022context} & \cmark & \cmark & \cmark &  &  & \cmark &  &  &  \\
TransPatch \cite{wang2022transpatch} &  & \cmark &  &  &  &  &  &  & \\
ZQA \cite{cai2022zqa} & \cmark & \cmark & \cmark &  &  & \cmark &  &  & \cmark\\
EBDA $^*$ \cite{cai2023ensemble} & \cmark & \cmark & \cmark &  &  & \cmark &  &  & \\
Shapira et al. \cite{shapira2023phantom} & \cmark &  & \cmark &  &  &  &  & \cmark & \\
\midrule[0.3pt]\bottomrule[1pt]
\end{tabular}
\end{adjustbox}

\end{table*}

\section{Evaluation Metrics on Adversarial Attacks and Model Robustness}
\label{sec: evaluation metrics}

We summarize evaluations metrics on adversarial attacks and model robustness in Figure~\ref{fig:taxonomy_od}. We include metrics used in existing works to evaluate attacks on object detection models, as well as propose relevant metrics from other domains, and elaborate on these metrics in Section~\ref{subsec: attack strength} to \ref{subsec: model robustness}. We summarize datasets and object detection models used in existing works in Section~\ref{subsec: datasets and models}.

\subsection{Attack strength}
\label{subsec: attack strength}

\subsubsection{Base model task performance}

Given a fixed Intersection-over-Union (IoU) threshold, a true positive is a predicted bounding box that has an IoU with a ground-truth box that meets or exceeds the threshold. A true negative is an object that the detector does not detect and is not in the ground-truth set. A false positive is a falsely detected object and a false negative is an object that the detector failed to detect.

\textbf{False positive rate (FPR)}: $FPR = \frac{\text{false positives}}{\text{false positives + true negatives}}$ is the ratio of falsely detected objects to the total number of negative examples.

\textbf{False negative rate (FNR)}: $FNR = \frac{\text{false negatives}}{\text{false negatives + true positives}}$ is the ratio of missed objects to the total number of objects.

The Average Precision (AP) and Mean Average Precision (mAP) are common metric for evaluating the performance of object detection algorithms. The definitions of the metrics differ in some cases. For clarify, we define the metrics below.

\textbf{Average Precision (AP@$\gamma$)}: We define AP@$\gamma$ as the precision averaged across all object classes at a fixed IoU threshold $\gamma$.

\textbf{Mean Average Precision (mAP)}: We define mAP as the average of AP@$\gamma$ across $T$ IoU thresholds $\{\gamma_t\}_{t=1}^T$, i.e. $mAP = \frac{1}{T} \sum_{t=1}^T \text{AP@}\gamma_t$. mAP consolidates detection performance at different IoU thresholds, and thus avoids having to set a fixed threshold for evaluation.

\subsubsection{Attack effectiveness}

\textbf{Change in task performance}: The magnitude of decrease in mAP from the base level reflects the overall effectiveness of the adversarial attacks. An increase in FPR indicates successful attacks for miscategorization and object fabrication. An increase in FNR indicates successful attacks for miscategorization and object vanishing.

\textbf{Fooling ratio}: Ratio of successfully attacked images. The success of an attack needs to be defined.
For instance, for object vanishing attacks, \cite{Wu2020DPAttackDP} define a successful attack as one which suppresses all bounding boxes in an image.

For attacks aimed at object vanishing, \cite{li2021udos} further utilizes image-level and instance-level blind degree to measure the effectiveness of the attacks.

\textbf{Image-level blind degree}: Ratio of images where at least one object is detected with confidence above a specified threshold.

\textbf{Instance-level blind degree}: Average number of objects detected with confidence above a specified threshold in each image.

With similar metrics, Wu \textit{et al.} proposed to compute the ratio between the metrics evaluated on the original and perturbed images~\cite{Wu2020DPAttackDP}.

\subsection{Time cost}

The frame rate for real-time object detection depends on the application. Detecting fast objects as in traffic monitoring requires at least 10 frames per second (FPS), while detecting slow objects such as monitoring people passing through an area requires 2-3 FPS \cite{lee2021realtime}. Attack methods that take a long time to generate perturbations are not suitable for real-time object detection applications. From the evaluation in \cite{xu2020evaluation}, UEA \cite{wei2019uea} and \cite{thys2019surveillance} take at most 0.05s per image and can be suitable for real-time attacks, while DAG \cite{xie2017dag}, Daedalus \cite{wang2022daedalus}, R-AP \cite{li2018rap} and \cite{yang2018cloak} take from 1 to more than 10 minutes.

\subsection{Number of queries}

Attack methods may repeatedly query the target model to access parameter gradients for optimization and to check for attack success. For example in \cite{xu2020evaluation}, DAG \cite{xie2017dag} and R-AP \cite{li2018rap} use between 50 and 100 queries, while Daedalus \cite{wang2022daedalus} uses more than 600 queries. When the adversary does not have the ability to query the target model a large number of times, these query-based methods would be unsuitable. Moreover, by querying the target model, the adversary risks being discovered. 

\subsection{Perturbation distortion}

\subsubsection{Perception}

An adversarial attack can be categorized as either perceptible or imperceptible based on whether the perturbation can be easily perceived by humans. In general, patch attacks are more easily perceived, while pixel perturbations regularized by a norm constraint are more difficult to perceive.

\subsubsection{Measurement}

The amount of distortion on the original image can be quantitatively assessed. Common measurements are $L_p$-norm with $p\in\{0,1,2,\infty\}$, proportion of perturbed pixels, Peak Signal-to-Noise Ratio (PSNR) as in Equation~\ref{eqn: psnr}, and Structural Similarity Index Measure (SSIM) comparing the luminance, contrast and structure between two images \cite{zhou2004ssim}.

\begin{equation}
    \begin{aligned}[b]
    & SSIM(x, x')=\\
    & luminance(x, x')^\alpha \cdot contract(x, x')^\beta \cdot structure(x, x')^\gamma \label{eqn: ssim}
    \end{aligned}
\end{equation}

\begin{align} 
    luminance(x, x') &= \frac{2\mu_x\mu_{x'} + k_l}{\mu_x^2 + \mu_{x'}^2 + k_l} \nonumber \\
    contrast(x, x') &= \frac{2\sigma_x\sigma_{x'} + k_c}{\sigma_x^2 + \sigma_{x'}^2 + k_c} \nonumber \\
    structure(x, x') &= \frac{\sigma_{x,x'} + k_s}{\sigma_x\sigma_{x'} + k_s} \nonumber
\end{align}

\noindent with constants $k_l, k_c, k_s$. The terms of $\mu_x$ and $\sigma_x^2$ are the pixel mean and variance of $x$ (and similarly defined for $x'$), and $\sigma_{x,x'}$ is the covariance of $x$ and $x'$.

\subsection{Transferability}
\label{subsec: transferability}

\begin{table*}[ht]

\caption{Models used in literature to evaluate adversarial attacks in object detection. \protect\label{tab: models_od}}

\centering
\begin{adjustbox}{max width=0.8\textwidth}
\begin{tabular}{lll}
\toprule[1pt]\midrule[0.3pt]

\textbf{Detector Type} & \textbf{Model / Algorithm} & \textbf{Version / Backbone}\\ \midrule
\multirow{ 4}{*}{One-stage Anchor-based}  & SSD & SSD300 VGG-16, SSD512 VGG-16 \\ 
                                        & YOLOv2 & Darknet-19\\
                                        & YOLOv3 & Darknet-53\\
                                        & YOLOv4 & CSPDarknet-53\\
                                        & RetinaNet & ResNet-101 \\ 
                                        & EfficientDet & EfficientNet + BiFPN \\ \hdashline
\multirow{ 6}{*}{Two-stage Anchor-based}  & R-FCN & ResNet-101 \\
                                        & Faster R-CNN & ZFNet, VGG-16, ResNet-50, ResNet-101, ResNeXt-101 \\
                                        & Libra R-CNN & ResNet-50 \\
                                        & Mask R-CNN & ResNeXt-101 \\
                                        & Cascade R-CNN & ResNet-101\\
                                        & Cascade Mask R-CNN & ResNeXt-101 \\
                                        & Hybrid Task Cascade & ResNeXt-101 \\ \hdashline
\multirow{ 2}{*}{One-stage Anchor-free} & FCOS & ResNet-50 \\  
                                        & FoveaBox & ResNet-50 \\
                                        & FreeAnchor & ResNet-50\\
                                        & DETR & DETR ResNet-50, Deformable DETR ResNet-50 \\    
\midrule[0.3pt]\bottomrule[1pt]
\end{tabular}
\end{adjustbox}

\end{table*}

Attack transferability is evaluated by computing the adversarial perturbations for an input image on a training model, and applying the perturbations to attack a different image or target model. Attacks that demonstrate high transferability to target models which differ from the training model can be suitable for gray-box and black-box settings.

\textbf{Cross-image}: Perturbation computed on an input image is applied on a different image. 

\textbf{Cross-resolution}: Target model processes images at a different resolution.

\textbf{Cross-training-dataset}: Target model is trained with a different dataset.

\textbf{Cross-backbone}: Target model has a different network architecture. Other aspects of the target model are assumed to be the same as those of the training model.

\textbf{Cross-model}: Target model is a different algorithm from the training model. For instance, training and target models are both one-stage detectors but different algorithms (e.g. SSD $\rightarrow$ YOLO), or
training model is a one-stage detector and target model is a two-stage detector (e.g. SSD $\rightarrow$ Faster R-CNN). Cross-model transferability can be difficult to achieve as models may be trained differently and hence use different features for predictions. Learning the perturbation on an ensemble of models have been found to increase attack effectiveness on the unseen target model \cite{wang2022daedalus, shapira2023phantom, cai2023ensemble}.

\subsection{Model Robustness}
\label{subsec: model robustness}

Let $\textit{perf}_{clean}(f)$ and $\textit{perf}_{adv}(f)$ be the task performance of model $f$ on clean and adversarially perturbed images, respectively. With multiple adversarial methods $\{adv_i\}_i$ and corresponding task performances $\{\textit{perf}_{adv_i}(f)\}_i$, we can obtain an overall performance metric as the mean or worst-case performance.

\textbf{Robust task performance:} Model performance $\textit{perf}_{adv}(f)$ on attacked images. Multiple models are ranked by $\textit{perf}_{adv}(f)$ for comparison.

To evaluate model robustness with respect to a baseline, \cite{taori2020measuring} proposes effectiveness and relative robustness metrics. These metrics are originally proposed for natural distribution shifts in image classification, but can be readily adapted to our task with our definitions of $\textit{perf}_{clean}(f)$ and $\textit{perf}_{adv}(f)$. 

\textbf{Effective robustness:} Let $\beta(\textit{perf}_{clean}(f))$ be the baseline performance on clean images.  \cite{taori2020measuring} instantiates $\beta$ as a log-linear function on $\textit{perf}_{clean}(f)$ across models $\{f_m\}_m$. Then the effective robustness of a model $f$ is:
\begin{equation}
    \rho_{eff}(f) = \textit{perf}_{adv}(f) - \beta(\textit{perf}_{clean}(f)).
\end{equation}
Models with special robustness properties will have $\textit{perf}_{adv}(f))$ above the $\beta$ fit.

\textbf{Relative robustness:} To compare $f_1$ with $f_2$, relative robustness is computed as:
\begin{equation}
    \rho_{rel}(f) = \textit{perf}_{adv}(f_1) - \textit{perf}_{adv}(f_2).
\end{equation}
$f_1$ and $f_2$ can be two different models (for example, one of them can be a baseline model), or two versions of a model with and without robustness intervention.

\subsection{Datasets and Models}
\label{subsec: datasets and models}

We summarize the datasets and computer vision models used to evaluate adversarial attacks for object detection in existing literature in Tables~\ref{tab: models_od} and \ref{tab: datasets_od}, respectively. The common datasets used are PASCAL VOC \cite{everingham2010voc} and MS COCO \cite{Lin2014MicrosoftCC}. The common models tested are anchor-based detectors such as SSD \cite{Liu2015SSD}, YOLO v2-v4 \cite{redmon2017yolov2, Redmon2018YOLOv3, Bochkovskiy2020YOLOv4} and Faster R-CNN \cite{Ren2015FasterRCNN}.

\begin{table}[htb]

\caption{Datasets used in literature to evaluate adversarial attacks for object detection. $\dagger$ denotes that dataset requires additional annotation. \protect\label{tab: datasets_od}}

\centering
\begin{adjustbox}{max width=\columnwidth}
\begin{tabular}{lll}
\toprule[1pt]\midrule[0.3pt]

\textbf{Data Type} & \textbf{Classes} & \textbf{Dataset}\\ \midrule
\multirow{ 2}{*}{Image} & Common objects & VOC-2007, VOC-2012, COCO, KITTI\\
                        & Person & Inria Person, CityPersons, ImageNet$^\dagger$ \\ 
                        & Traffic sign & Stop Sign, Mapilliary$^\dagger$ \\ \hdashline
Video                   & Common objects & ImageNet VID \\
\midrule[0.3pt]\bottomrule[1pt]
\end{tabular}
\end{adjustbox}

\end{table}
\section{Defenses against Adversarial Attacks}
\label{sec: defenses}

Extensive research efforts have investigated defense mechanisms against adversarial attacks in image classification, and we refer readers to existing survey papers \cite{costa2024survey, khamaiseh2022survey} for a comprehensive overview. The defensive strategies include adversarial training, knowledge distillation, gradient regularization, network architecture search for robust model designs, auxiliary detector networks for adversarial example identification, denoising adversarial inputs and feature representations, and adversarial purification through generative models to mitigate the impact of perturbations.

In comparison, literature on defenses in object detection is more limited. Dong \textit{et al.}~\cite{amirkhani2022surveyav} reviewed the defenses focused on applications in autonomous vehicles. In general, defense strategies can be categorized into three primary groups: 1) training procedure modifications, 2) input/feature denoising, and 3) external attack detection networks. In the first group, Dong \textit{et al.}~\cite{dong2022robustdet} improved the training procedure to disentangle model gradients on clean and adversarial samples during training, while Saha \textit{et al.}~\cite{saha2020spatial} developed methods to limit the dependence on spatial context during training. Chen \textit{et al.}~\cite{chen2021classaware} introduced class-wise loss normalization to balance the influence of each object class, and Zhang and Wang~\cite{zhang2019robust} proposed to generate attack samples using multiple loss components for adversarial training. Chiang \textit{et al.}~\cite{chiang2020certified} developed the method to certify object detection predictions by aggregating multiple predictions for randomly perturbed inputs via median smoothing.
For the input denoising approach, Zhou \textit{et al.}~\cite{zhou2024preprocessing} developed the method to transform the input images to remove non-robust features and adversarial noise. To defend against patch attacks, Liu \textit{et al.}~\cite{liu2022segment} introduced the method to localize adversarial patches and remove them from the images, while Kim \textit{et al.}~\cite{kim2022energy} presented the approach to eliminate adversarial features through learned detection mechanisms. The third group comprises detection-based defenses that aim to identify adversarial samples and trigger appropriate alerts. For instance, Strack \textit{et al.}~\cite{Strack2023infrared} and \textit{et al.}~\cite{Ji2021AdversarialYolo} focused on training samples with randomly added patches to detect adversarial patches. Xiang and Mittal \textit{et al.}~\cite{xiang2021detectorguard} ultilized an image classifier and object detector to validate prediction objectness, where predictions lacking sufficient explanatory support are flagged as potential attacks.

\section{Evaluation}
\label{sec: evaluation}

This section thoroughly evaluates the open-sourced adversarial attacks listed in Table \ref{tab: methods_literature_od}. Our experiments aim to compare available open-source attack methods and provide a comprehensive analysis of their performances across various object detector architectures.

Specifically, section \ref{sssec: attack_effectiveness} evaluates and compares the attack effectiveness of 5 open-source attack methods, including Dpatch \cite{Liu2019DPATCH}, TOG \cite{chow2020tog}, DPAttack \cite{Wu2020DPAttackDP}, CAT \cite{cai2022context}, EBAD \cite{cai2023ensemble}. For TOG \cite{chow2020tog}, we reimplemented the algorithms from TensorFlow to PyTorch to evaluate them with MMDetection’s pretrained object detectors, in a manner consistent with other attacks in our experiment. For DPatch \cite{Liu2019DPATCH}, we used the reimplementation from Adversarial Robustness Toolbox (ART), an open-source Python library for Machine Learning Security \footnote{https://github.com/Trusted-AI/adversarial-robustness-toolbox}. Other methods were excluded due to outdated or unmaintained code, or the absence of a well-specified environment and setup for rerunning. We evaluate these attacks on Faster R-CNN and YOLOv3, two object detectors that are commonly used in literature.
Section \ref{sssec: ensemble_att} focuses on studying cross-model black-box attacks on 7 different object detectors with 49 surrogate-victim pairs. We highlight the effects of different detector architectures on the attack performance and their robustness against the attacks. Lastly, section \ref{sssec: transfer} provides a transferability study of ensemble-based attacks on 26 object detectors, ranging from traditional to modern state-of-the-art ones, including detectors with vision-language pretraining which have not been thoroughly studied in existing literature.

\subsection{Experiment Setup} \label{ssec: experiment_setup}

For a fair comparison, we standardized attack hyperparameters including the maximum number of queries and perturbation budget, and ran all attacks on a single GeForce GTX 1080 12GB RAM. 



\textbf{Dataset}: We conduct evaluations in sections \ref{sssec: attack_effectiveness} and \ref{sssec: transfer} using the COCO 2017 validation set with 5000 images on 80 object classes. For the evaluation on ensemble attacks in section \ref{sssec: ensemble_att}, we use a subset of 500 random images (10\%) from the COCO 2017 validation set to evaluate 49 surrogate-victim pairs. Using a smaller subset helps to reduce the time and cost of running this evaluation.

\textbf{Object detectors}: We evaluate attacks on object detectors commonly used in literature: Faster-RCNN with a ResNet-50 backbone (denot. FR-RN50) and YOLOv3 with a Darknet-53 backbone (denot. YOLOv3-D53). We also include other object detectors, such as RetinaNet, Libra R-CNN and FCOS, and modern state-of-the-art detectors such as DeTR with Transformer architecture, Grounding DINO, and GLIP with vision-language pretraining. 

We point out that different frameworks (Torch, MMDetection, Tensorflow) may provide different model pretrained weights, and these weights may also be periodically updated. Thus, even using the same detector architecture may yield mAP scores different from those originally reported in the enlisted papers in Table \ref{tab: methods_literature_od}. For a fair comparison, we use the pretrained model weights provided by MMDetection v3.3.0 pretrained on COCO 2017 train set for all evaluations. Their mAP on clean images are presented in Table~\ref{tab: clean_map}.




\begin{table}[htb]

\caption{The mAP score of FR-RN50 and YOLOv3 evaluated on the clean COCO 2017 validation set.
\protect\label{tab: clean_map}}

\centering
\begin{adjustbox}{max width=\columnwidth}
\begin{tabular}{lcc}
\toprule[1pt]\midrule[0.3pt]

Model & Dataset & mAP Clean (\%)\\
\midrule

FR-RN50 & COCO 2017 & 58.10\\
YOLOv3-D53 & COCO 2017 & 52.80\\

\midrule[0.3pt]\bottomrule[1pt]
\end{tabular}
\end{adjustbox}


\end{table}

\textbf{Maximum queries and perturbation budget}: 
We refer to \textit{query} as the number of times the attack algorithm retrieves the detection result from the victim detector to update its perturbation. Note that this is typically called an \textit{iteration} in white-box settings. By default, we limit the maximum number of queries to 10. Several works allowed as many as 5,000 - 100,000 queries, which is impractical as repeated queries may be easily detected by defense mechanisms. We also reported attack performances under a single query run. We set $L_{\infty}=10/255$, a typical value set in literature. 

\textbf{Metrics}: We calculate the mAP of the attacked images and compare it to the mAP of the clean images. Given that different versions of pretrained object detectors may exhibit different mAP scores on clean images, we provide the \textit{percentage drop in mAP} as a normalized metric for easy comparison, calculated by $(1 - \frac{mAP \  Adv.}{mAP \ Clean})*100\%$. We also report the \textit{time cost}, defined as the time (in seconds) needed to generate the perturbations, including the iterative updating process.

Note that not all adversarial attacks aim to reduce the mAP score, thus their observed mAP drops might not appear substantial. Also, while \textit{fooling rate} is a popular metric, we do not include it in our evaluation. This is because each work defines the criteria for successful attacks differently based on their specific setup and objectives, making the fooling rate non-comparable across different attacks. 


\subsection{Results and Analysis}

\subsubsection{Attack effectiveness} \label{sssec: attack_effectiveness}


\begin{table*}[htb]

\caption{Results of adversarial attack on FR-RN50. $L_{\infty}$ denotes whether the attack has constrained  $L_{\infty}=10/255$. For the patch attack DPatch, the pixel change can vary from 0.0 to 255.0. Attacks with \textit{ensemble} indicates that these methods use a group of two surrogates FR-RN50 and YOLOv3-D53. EBAD single is the EBAD attack with a single surrogate FR-RN50. The sign $^*$ denotes reimplemented methods. \protect\label{tab: eval_tab_fr}}

\centering
\begin{adjustbox}{max width=\textwidth}
\begin{tabular}{lccccccccc}
\toprule[1pt]\midrule[0.3pt]

\multicolumn{10}{c}{FR-RN50 victim}\\
\midrule
\textbf{Attack name} & $L_{\infty}$ & \multicolumn{4}{c}{\textbf{10-query}} & \multicolumn{4}{c}{\textbf{Single query}} \\
\cmidrule(lr){3-6}    
\cmidrule(lr){7-10}

&  & mAP Drop (\%) & mAP Adv. (\%) & Time (s) & PSNR & mAP Drop (\%)  & mAP Adv. (\%) & Time (s) & PSNR\\
\midrule
DPatch $^*$ & \xmark & 6.37 & 54.40 & 4.51  & 36.07 & 4.82  & 55.30 & 0.21 & 36.21\\
TOG untargeted $^*$ & \cmark & 55.59 & 25.80  & 3.86  & 32.13 &  16.87 & 48.30 & 0.80 & 32.79\\
TOG fabricated $^*$ & \cmark & 14.77 & 45.00  & 8.66  & 32.02 & 11.53 & 51.40& 1.12 & 32.76\\
TOG vanishing $^*$ &  \cmark & 30.80 & 40.20  & 6.94 & 32.20 & 22.89 & 44.80 & 0.76 & 32.78 \\
DPAttack & \xmark & 19.96 & 46.50 & 2.37 & 25.57 & 19.44 & 46.80 & 0.33 & 25.62\\
CAT ensemble & \cmark & 16.69 & 48.40 & 28.55 & 33.26 & 6.37 & 54.40 & 4.18 &33.31 \\
EBAD single & \cmark & 22.37 & 45.10 & 20.03 & 31.83 & 24.26 & 44.00 & 5.10 & 32.05\\
EBAD ensemble & \cmark & 22.55 & 45.00 & 32.32 & 31.76 & 23.58 & 44.40 & 3.75 & 31.96\\

\midrule[0.3pt]\bottomrule[1pt]
\end{tabular}
\end{adjustbox}

\end{table*}

\begin{table*}[htb]

\caption{Results of adversarial attack on FR-RN50. $L_{\infty}$ denotes whether the attack has constrained  $L_{\infty}=10/255$. For the patch attack DPatch, the pixel change can vary from 0.0 to 255.0. Attacks with \textit{ensemble} indicates that these methods use a group of two surrogates FR-RN50 and YOLOv3-D53. EBAD single is the EBAD attack with a single surrogate YOLOv3-D53. The sign $^*$ denotes reimplemented methods. \protect\label{tab: eval_tab_yolov3}}

\centering
\begin{adjustbox}{max width=\textwidth}
\begin{tabular}{lccccccccc}
\toprule[1pt]\midrule[0.3pt]

\multicolumn{10}{c}{YOLOv3-D53 victim}\\
\midrule
\textbf{Article} & $L_{\infty}$ & \multicolumn{4}{c}{\textbf{10-query}} & \multicolumn{4}{c}{\textbf{Single query}}\\
\cmidrule(lr){3-6}    
\cmidrule(lr){7-10}

& & mAP Drop (\%) & mAP Adv. (\%) & Time (s) & PSNR & mAP Drop (\%)  & mAP Adv. (\%) & Time (s) & PSNR\\
\midrule
DPatch $^*$ & \xmark & 3.97 & 50.70  & 1.04 & 35.18 & 2.65 & 51.40 & 0.12 & 36.04\\
TOG untargeted $^*$ & \cmark & 60.22 & 21.70  & 0.85  & 32.14 & 14.77 & 45.00 & 0.31 & 32.71\\
TOG fabricated $^*$ & \cmark & 27.46 & 38.30 & 0.85  & 32.19 & 13.63 & 45.60 & 0.31 & 32.71 \\
TOG vanishing $^*$ & \cmark & 39.20 & 32.10  & 1.16 & 32.23 & 24.43 & 39.90 & 0.16 & 32.76 \\
DPAttack & \xmark & 22.68 & 45.00 & 2.37 & 25.57 & 22.68 & 45.00 & 0.33 & 25.62\\
CAT ensemble & \cmark & 16.09 & 44.30 & 28.55 & 33.26 & 8.90 & 48.10 & 4.18 & 33.31\\
EBAD single & \cmark & 21.97 & 41.20 & 7.65 & 31.79 & 25.75 & 39.20 & 1.47 & 32.07\\
EBAD ensemble & \cmark & 19.70 & 42.40  &  27.27 & 31.71 & 19.69 & 42.40 & 3.36 & 32.01\\

\midrule[0.3pt]\bottomrule[1pt]
\end{tabular}
\end{adjustbox}

\end{table*}

Tables \ref{tab: eval_tab_fr} and \ref{tab: eval_tab_yolov3} show the attack performances on victims YOLOv3-D53 and FR-RN50 respectively. 

\textbf{TOG is an effective white-box attack with low time cost:} White-box setting assumes access to all detection information, such as prediction scores, losses, and bounding box coordinates. Thus, the resulting mAP drops in this setup are often more significant compared to gray-box and black-box attacks. Our experiment finds that TOG-untargeted is an effective white-box attack that achieves mAP drops of more than 50\% with only 10 queries. In our evaluation, the TOG untargeted attack results in the highest mAP drops of 55.59\% on FR-RN50 and 60.22\% on YOLOv3-D53. It also has the lowest time cost compared to other attacks, with less than 4 seconds to attack FR-RN50 and less than 1 second to attack YOLOv3-D53. Older methods such as DAG \cite{xie2017dag} or UEA \cite{wei2019uea} require more than 30 queries yet still achieve poorer results \cite{chow2020understanding}.  

For patch attacks under white-box setting, DPatch is shown to be ineffective under our experiment setting with an insignificant mAP drop of only 6.37 \%. DPatch requires a large number of queries (iterations), between 1000 and 20000 \cite{Liu2019DPATCH}, which explains why it falls short in our experiment with a limited but more realistic 10 queries. Its attack success is also sensitive to the location of the patch. If the patch is placed where objects are present in the image, the chance of success increases. On the other hand, DPattack produces a significant mAP drop of 22.68 \%. Nevertheless, a common issue with patch attacks is that the perturbed patch added to the image is easily recognizable by human eyes, making them easy to be detected in real-world scenarios.


EBAD is specifically designed for ensemble-based black-box attacks, yet still shows a significant mAP drop in the white-box setting with a single surrogate. In Tables \ref{tab: eval_tab_fr} and \ref{tab: eval_tab_yolov3}, EBAD-single refers to using the same surrogate model (FR-RN50 or YOLOv3-D53) to attack the same victim architecture, achieving mAP drops of 22.37\% on FR-RN50 and 21.97\% on YOLOv3-D53. 

For the ensemble-based CAT and EBAD, we use a group of two surrogates FR-RN50 and YOLOv3, and separately test the results on 2 victim detectors FR-RN50 and YOLOv3-D53. CAT, a transfer attack method that generates a universal perturbation from surrogates to attack unseen victim detectors, achieves approximately 16\% mAP drops. Our experiment shows that EBAD-ensemble achieves impressive mAP drops of more than 22.55\% within 10 queries. Moreover, its attack performance with a single query also outperforms all other methods. We also notice that the mAP drops from EBAD's single query attacks are slightly higher than those from the 10-query attacks. This could potentially be due to their approach of adapting the PGD \cite{madry2018towards} optimization from white-box adversarial attacks for image classification, which aims to reduce accuracy scores rather than directly targeting mAP scores. From what we observed,  within the first iterations of PGD, the average perturbation values were typically at the maximum $L_{\infty}=10$ constraint. However, after several iterations, they were decreased to around 5, making the perturbation less visible to the human eye while still focusing on accuracy reduction. This explains why for the EBAD, increasing the number of queries may not necessarily lead to higher mAP drops in our experiment.

Earlier black-box attacks such as PRFA \cite{liang2021prfa} and RAD \cite{chen2022rad} require several constraints to be successful. They both require knowing the detector architecture to adjust its optimization process. Moreover, RAD sets a larger perturbation norm $L_{\infty}=16/255$. Meanwhile, ensemble-based methods such as CAT \cite{cai2022context} and EBAD \cite{cai2023ensemble} use a group of surrogate detectors to attack another victim's black-box detector that requires no prior knowledge about the victim, which is a more practical use case.



\subsubsection{Cross-model black-box attack} \label{sssec: ensemble_att} In this experiment, we study the attack effictiveness of different surrogate-victim detector pairs in the black-box setting. Table \ref{tab: ensemble_matrix} evaluates cross-model black-box attack performance with different sets of attack and victim models ranging from traditional detectors (Faster R-CNN, YOLOv3, RetinaNet, Libra R-CNN, FCOS) to modern state-of-the-art detectors such as DeTR with Transformer architecture, and GLIP with vision-language pretraining. We conduct our experiment with EBAD attack due to its potential wider applicability as an imperceptible black-box attack. We use a subset of 500 samples from the COCO 2017 validation set for evaluation.


\textbf{Robustness against white-box and robustness against black-box attacks are not correlated.} We observe that stronger detectors, such as DeTR and GLIP, which achieve higher mAP on clean images also tend to be relatively robust against adversarial attacks. In particular, DeTR-RN50 and GLIP Swin-T have respective white-box mAP drops of 8.01\% and 14.85\%, the lowest amongst the models tested in Table~\ref{tab: ensemble_matrix}. Interestingly however, models that are vulnerable against white-box attacks can still be robust against black-box attacks. YOLOv3-MN has the highest white-box mAP drop of 43.38\% but the lowest black-box mAP drop of 4.41\%. DeTR-RN50 attacked by YOLOv3-MN has a mAP drop of 8.17\%, while YOLOv3-MN attacked by DeTR-RN50 has a lower mAP drop of 3.37\%.

\textbf{Using a surrogate with the same backbone architecture with the victim does not necessarily result in a high mAP drop}. In Table~\ref{tab: ensemble_matrix}, 5 models use a ResNet-50 backbone. We observe that attacking a victim model with a surrogate that shares the backbone architecture generally results in comparatively higher mAP drop, but the value of mAP drop is not consistently high. For instance, RetinaNet-RN50 have mAP drops above 10\% (and up to 19.05\%) when attacked by surrogates with ResNet-50 backbone, while FCOS-RN50 and DeTR-RN50 have no mAP drops above 10\%.


\begin{table*}[htb]

\caption{The mAP score drop (\%) of EBAD attack with different surrogate and victim detectors. Entries on the diagonal are equivalent to white-box attacks as they use the same detectors for the attacker and victim. The \textit{Mean mAP drop} column reports the mean mAP drop for each victim with different surrogates. \protect\label{tab: ensemble_matrix}}

\centering
\begin{adjustbox}{max width=\textwidth}
\begin{tabular}{lccccccccc}
\toprule[1pt]\midrule[0.3pt]
\multirow{2}{*}{\diagbox{\textbf{Victim}}{\textbf{Surrogate}}} & \textbf{FR} & \textbf{YOLOv3} & \textbf{YOLOv3} & \textbf{RetinaNet} & \textbf{FCOS} & \textbf{DeTR} & \textbf{GLIP} & \textbf{Mean} & \textbf{Mean black-box}\\

& \textbf{-RN50} & \textbf{-D53} & \textbf{-MN} & \textbf{-RN50} & \textbf{-RN50} & \textbf{-RN50}& \textbf{Swin-T} & \textbf{mAP drop} & \textbf{mAP drop}\\
\cmidrule(lr){1-8}
\cmidrule(lr){9-9}
\cmidrule(lr){10-10}

FR-RN50 & \textbf{17.44} & 10.19 & 7.77 & 13.99 & 10.02 & 8.98 & 8.81 & 11.03 & 9.96\\
YOLOv3-D53 & 6.03 & \textbf{22.83} & 6.22 &3.77 & 4.72 & 4.90 & 4.90 & 7.63 & 5.10\\
YOLOv3-MN & 3.79 & 7.58 & \textbf{43.38} & 2.95 & 4.63 & 3.37 & 4.21 & 9.98 & 4.41\\
RetinaNet-RN50 & 19.05 & 10.34 & 7.10& \textbf{27.40} & 10.52 & 11.25 & 8.35 & 13.43 & 11.10\\
FCOS-RN50 & 6.57 & 6.07 & 6.56 & 7.72 & \textbf{18.27} & 6.90 & 5.91 & 8.28 & 6.62\\
DeTR-RN50 & 7.37 & 8.01 & 8.17 & 8.81 & 7.21 & \textbf{8.01} & 7.53 & 7.87 & 7.85\\
GLIP Swin-T & 4.90 & 5.17 & 4.90 & 4.22 & 5.31& 3.00& \textbf{14.85} & 6.05 & 4.58\\
\midrule[0.3pt]\bottomrule[1pt]
\end{tabular}
\end{adjustbox}

\end{table*}


\subsubsection{Transferability study}\label{sssec: transfer} 

In this experiment, we evaluate the transferability of the ensemble-based attack EBAD on different unseen detectors. Among those, Grounding DINO \cite{liu2023gdino} and GLIP \cite{li2021glip} are state-of-the-art detectors that achieve remarkably high detection results on various benchmark datasets, including COCO 2017.

We conduct this experiment on the full COCO 2017 validation set, which includes 5000 samples. We assess cross-model transferability using 4 sets of perturbed images, generated from different surrogate groups as follows:
\begin{itemize}
\item Images perturbed by 2 surrogates FR-RN50 and YOLOv3-D53, using RetinaNet-RN50 as the victim.
\item Images perturbed by 4 surrogates: one-stage anchor-based models (YOLOv3-D53, SSD), a two-stage anchor-based model (FR-RN50), and a one-stage anchor-free model (FCOS), with RetinaNet-RN50 as the victim.
\item Images perturbed by 2 surrogates YOLOv3-D53 and GLIP Swin-T, with RetinaNet-RN50 as the victim.
\item Images with random noise of $L_{\infty}=10$, provided as a baseline for comparison.
\end{itemize}

The output images from these sets of attacks on RetinaNet-RN50 are then tested on 26 unseen detectors. In Table \ref{tab: transfer}, we report the \% mAP drops of these unseen detectors for different object areas, following the criteria from MMDetection:

\begin{itemize}
\item Small objects: Areas in an image are less than 1024 (32x32 pxiels)
\item Medium objects: Areas in an image are from 1024 to 9216 (32x32 to 96x96 pixels)
\item Large objects: Areas in an image greater than 9216 (96x96 pixels)
\end{itemize}


All of the object detectors in this evaluation are pretrained on COCO 2017.

\begin{table*}[htb]

\caption{The mAP score drop (\%) of EBAD attack on unseen victim detectors. \textit{Transfer} distinguishes evaluation on unseen victims (\cmark) with seen victims (\xmark). \textit{All} reports the normal \% mAP drops at IoU 0.5 for all-sized objects. \textit{Small, Medium, Large} reports the \% mAP drops at IoU 0.5:0.95 on small, medium, and large objects respectively. \textit{Noise} reports the \% mAP drop from random noise.
\protect\label{tab: transfer}}

\centering
\begin{adjustbox}{max width=\textwidth}
\begin{tabular}{lllccccccccccccccc}
\toprule[1pt]\midrule[0.3pt]
\textbf{Victim model} & \textbf{Backbone} & \textbf{Noise} & \multicolumn{5}{c}{\textbf{FR-RN50} + \textbf{YOLOv3-D53}} & \multicolumn{5}{c}{\textbf{FR-RN50 + YOLOv3-D53 + FCOS-RN50 + SSD}} & 
\multicolumn{5}{c}{\textbf{YOLOv3-D53 + GLIP Swin-T}}\\
\cmidrule(lr){3-3}
\cmidrule(lr){4-8}
\cmidrule(lr){9-13}
\cmidrule(lr){14-18}
& & All & Transfer & All & Small & Medium & Large & Transfer & All & Small & Medium & Large & Transfer & All & Small & Medium & Large\\
\midrule

\multirow{3}{*}{Faster R-CNN} & Resnet-50 & 5.85 & \xmark & 27.71 & 37.26 & 27.32 & 27.65 & \xmark & 35.63 & 41.03 & 34.87 & 33.13
& \cmark & 13.77 & 24.53	& 15.37	& 11.23
\\
& Resnet-101 & 5.32 & \cmark & 15.14 & 27.23 & 16.93 & 13.50 & \cmark & 23.29 & 33.04 & 25.86 & 20.94
& \cmark & 12.15 & 27.23 & 13.73 & 9.39
\\
& ResneXt-101 & 5.31 & \cmark & 14.49 & 25.00 & 15.38 & 12.71 & \cmark & 22.70 & 31.25 & 24.83 & 21.49
& \cmark & 12.08 & 24.17 & 14.07 & 10.28
\\
\midrule

\multirow{2}{*}{YOLOv3} & Darknet-53 & 2.84 & \xmark & 21.59 & 20.83 & 17.66 & 22.37 & \xmark & 25.38 & 23.61 & 22.15 & 28.41
& \xmark & 16.10 & 16.67 & 12.87 & 15.66
\\
& MobileNet-V2 & 2.41 & \cmark & 7.03 & 15.09 & 6.77 & 6.57 & \cmark & 9.45 & 15.10 & 8.76 & 8.28
& \cmark & 5.93	& 4.72 & 5.58 & 4.86
\\
\midrule

\multirow{2}{*}{RetinaNet} & Resnet-50 & 5.78 & \xmark & 23.29& 33.82  & 23.57 & 23.49 &  \xmark & 31.95 & 38.23 & 32.75 & 33.89
& \cmark & 14.62 & 25.98 & 16.13 & 13.51
\\
& Resnet-101 & 5.20 & \cmark & 15.45 & 28.00 & 16.35 & 12.87 & \cmark & 23.09 & 35.02 & 24.77 & 21.58
& \cmark & 12.50 & 26.27 & 13.55 & 10.50
\\
\midrule

\multirow{1}{*}{Libra R-CNN} & Resnet-50 & 5.21 & \cmark & 19.83 & 27.60 & 20.95 & 19.17 & \cmark & 27.56 & 33.48 & 29.28 & 26.60
& \cmark & 12.10 & 22.17 & 13.81 & 10.31
\\
\midrule

FCOS & Resnet-50 & 4.41 & \cmark & 11.76 & 20.81 & 13.94 & 11.11 & \xmark & 26.80 & 30.61 & 27.89 & 30.11
& \cmark & 9.80 & 20.00 & 12.42 & 9.32
\\
& Resnet-101 & 4.79 & \cmark & 14.70 & 25.78 & 16.05 & 14.17 & \cmark & 24.27 & 33.33 & 26.05 & 24.75
& \cmark & 11.45& 22.67& 13.49& 11.58
\\
& ResneXt-101 & 4.96 & \cmark & 13.76 & 26.15 & 15.05 & 12.43 & \cmark & 20.32 & 29.23 & 23.01 & 20.29
& \cmark & 11.20& 24.23& 13.98& 10.05
\\
\midrule

DeTR & Resnet-50 & 5.54 & \cmark & 16.61 & 27.98 & 17.61 & 17.52 & \cmark & 25.47 & 33.20 & 27.25 & 26.87
& \cmark & 12.50 & 24.63 & 13.42 & 11.90
\\
\midrule

RTMDet-T & - & 3.80 & \cmark & 8.46 & 12.38 & 9.67 & 8.23 & \cmark  & 12.26 & 16.67 & 13.63 & 12.86
& \cmark & 7.25 & 14.29 & 8.79 & 6.86
\\
RTMDet-M & - & 4.05 & \cmark & 9.75 & 19.21 & 9.61 & 7.07 & \cmark  & 11.39 & 22.47 & 12.94 & 10.22
& \cmark & 7.65 & 18.24 & 9.06 & 6.02
\\
RTMDet-L & - & 3.78 & \cmark & 7.41 & 17.06 & 9.25 & 5.40 & \cmark & 9.88 & 23.25 & 11.57 & 8.32
& \cmark & 6.98 & 18.53 & 8.72 & 4.96
\\
RTMDet-X & - & 3.83 & \cmark & 7.53 & 18.89 & 8.54 & 5.64 & \cmark & 9.80 & 19.17 & 11.15 & 7.22
& \cmark & 7.10 & 17.78 & 8.54 & 4.77
\\
\midrule

\multirow{5}{*}{GLIP} & swin-T (A) & 4.06 & \cmark & 9.10 & 18.16 & 10.71 & 9.02 & \cmark & 13.57 & 21.48 & 15.54 & 14.44
& \xmark & 12.17 & 20.72 & 14.16 & 11.73
\\
& swin-T (B) & 3.46 & \cmark & 7.33 & 16.20 & 8.38 & 6.79 & \cmark & 11.34 & 19.49 & 13.50 & 11.96
& \xmark & 9.96 & 20.51 & 11.97 & 9.45
\\
& swin-T (C) & 3.40 & \cmark & 6.80 & 15.16 & 8.18 & 6.37 & \cmark & 10.20 & 18.83 & 12.35 & 10.13 
& \xmark & 10.20 &  19.80 & 12.85 & 9.99
\\
& swin-T & 3.26 & \cmark  & 7.07 & 14.10 & 8.67 & 6.29 & \cmark  & 10.47 & 20.30 & 13.17 & 10.30
& \xmark & 12.93 & 22.03 & 15.33 & 13.73
\\
& swin-L & 1.94 & \cmark & 4.26 & 12.95 & 5.32 & 3.23 & \cmark & 6.33 & 16.07 & 8.92 & 4.31
& \cmark & 4.78 & 13.62 & 6.89 & 2.96
\\
\midrule

\multirow{2}{*}{GDINO} & swin-T & 2.91 &  \cmark  & 5.82 & 13.88 & 7.28 & 4.27 & \cmark  & 9.14 & 16.94 & 11.16 & 7.99
& \cmark & 7.95 & 16.94 & 9.55 & 6.20
\\
& swin-B & 2.45 & \cmark & 4.12 & 11.89 & 5.49 & 1.46 & \cmark & 6.06 & 14.19 & 8.00 & 3.46
& \cmark & 4.90 & 13.27 & 6.44&1.86
\\

\midrule[0.3pt]\bottomrule[1pt]
\end{tabular}
\end{adjustbox}

\end{table*}

\textbf{Detections on smaller objects are more vulnerable to adversarial attacks.} The mAP drops reported for small objects are significantly higher compared to medium or large objects when under attack. More traditional victim detectors, such as Faster R-CNN, YOLOv3, RetinaNet, Libra R-CNN, and FCOS, experience mAP drops for small objects of 10\%-20\% greater than those of the larger-sized objects. For newer victim detectors such as DeTR, RTMDet, GLIP, and Grounding DINO, the gaps are smaller yet remain significant. 


\textbf{Transferred attacks show decreased effects on more recent detector architectures.} There is a clear trend that the mAP scores decline less for more recent object detectors compared to older ones. More modern object detector may be more robust as they use novel architectures and are trained on larger data. RTMDet, GLIP, and Grounding DINO experience considerably smaller mAP drops, all below 10\%. Even when using GLIP Swin-T as the surrogate, the victims including GLIP itself, RTMDet, and Grounding DINO generally suffer smaller mAP drops. 

\textbf{A diverse group of surrogate detectors increases the adversarial attack transferability.} Utilizing an ensemble of 4 surrogate detectors results in a greater mAP drop across all unseen detectors. A broader range of surrogate models with diverse architectures and types can significantly increase the effectiveness of adversarial attacks. From Table \ref{tab: transfer}, we also show that the 4-surrogate group results in greater mAP drops than YOLOv3-D53 + GLIP Swin-T on new victim detectors such as RTMDet, GLIP, and Grounding DINO. 



\section{limitations}

While we made every effort to include all relevant adversarial attacks in our experiments, some methods were excluded due to outdated or unmaintained code, or the absence of a well-specified environment and setup for rerunning. For TOG \cite{chow2020tog}, we reimplemented the algorithms from TensorFlow to PyTorch to evaluate them with MMDetection’s pretrained object detectors, in a manner consistent with other attacks in our experiment. We reproduced the algorithm as best as we can based on the descriptions in the original manuscript \cite{chow2020tog}. For another work, DPatch \cite{Liu2019DPATCH}, we used the reimplementation from Adversarial Robustness Toolbox (ART), an open-source Python library for Machine Learning Security. Consequently, the results presented may differ from those reported in the original papers. Other methods were not included due to insufficient documentation on parameter setups. Despite attempts to contact the authors, we were unable to obtain the necessary information and code base to execute the methods. 

As mentioned in Section \ref{ssec: experiment_setup}, we standardized the number of queries and perturbation budget ($L_{\infty}$) to ensure comparability among different methods. As a result, methods such as DPatch \cite{Liu2019DPATCH} performed less effectively, as they were originally designed for larger query settings. The original implmentation of DPatch used up to 200k iterations, which is impractical especially for real time attacks. Moreover, variations in $L_{\infty}$ values can significantly impact the mAP outcome. Higher $L_{\infty}$ values generally result in a more substantial mAP drop, but they also make the perturbations more detectable to the human eye. Certain patch-based methods, such as DPatch and DPAttack, were easily visible to human observers and did not adhere to this $L_{\infty}$ constraint by design. Imposing a constrained $L_{\infty}$ on these methods nullified their attack effectiveness. Therefore, we did not set the constraint for these two methods in our experiment.




\section{Open areas for research}
\label{sec: open_areas}

\textbf{Adversarial attacks on modern object detectors have more room for research.} In our experiments, we observed that more recent object detectors, such as Transformer-based and vision-language pretrained ones, are potentially more robust against adversarial attacks. 
Modern object detectors may be more robust as they use novel architectures, training strategies and are trained on different or larger datasets. Prior research \cite{bai2021robust, bhojanapalli2021robust, daquan2022robust} focused on image recognition found that the robustness of transformers can be partially attributed to their training strategies and self-attention mechanisms. More research is needed on the architectural, training and dataset designs that significantly affect robustness for object detection.

\textbf{Defences on small objects are particularly needed.}
We observed that small objects within images are significantly more vulnerable to adversarial attacks. To our knowledge, there is currently no research specifically addressing adversarial attacks on small objects, either from the attack or defense perspective. We recommend further investigation into this issue in the future.

\textbf{Adversarial attacks based on image rendering are under-explored for the object detection task.}
Besides pixel value perturbations covered in Section~\ref{sec: taxonomy_od}, literature in image classification also explored attacks based on image rendering \cite{Engstrom2017ARA, Shamsabadi2019ColorFoolSA, duan2020natural}, including image rotation, translation, lighting, and coloring. That is, an adversarial attack is to render image $x$ such that the transformed image $x'=r(x)$ for a rendering function $r(\cdot)$ is incorrect i.e. $f(x';\Theta) = f(r(x);\Theta)\neq y$. 
Rendering-based attacks have practical implications in applications such as autonomous vehicles and surveillance systems, where adversarial rendering methods can simulate adverse weather conditions, camera angles and object orientations to test model robustness in constantly shifting environments.

\textbf{Development in multimodal models presents new areas for research on adversarial attacks and robustness.} For instance, applications such as autonomous driving and robotics require additional temporal and depth information. In these applications, it is common for models to integrate RGB images with other data sources such as LiDAR and 3D point clouds, and to process video data instead of a single image. These multimodal inputs offer richer contextual information, but also introduce new attack surfaces. For instance, adversaries could manipulate depth information from LiDAR or distort 3D point cloud data to mislead object detection systems, even when the RGB input appears unaffected. Similarly, perturbing temporal consistency in video data could degrade detection models that rely on motion cues. Research in this area is crucial to develop robust defense mechanisms capable of handling coordinated multimodal attacks.

\textbf{Physical adversarial attacks in object detection need benchmark evaluation.} Conducting evaluations in the physical world is often expensive and logistically challenging, particularly when it comes to automatically generating attacks for testing under various real-world conditions. The physical environment imposes constraints due to lighting conditions, occlusions, camera angles, and object distances, all of which can significantly affect the effectiveness of adversarial attacks and model robustness. While augmentations such as Expectation Over Transformation (EOT) have been employed to simulate real-world transformations, ensuring the fidelity of these simulated images to real-world scenarios remains a crucial challenge for accurate and reliable evaluation.

\section{Conclusion}
\label{sec: conclusion}

This paper provides a comprehensive survey and evaluation of adversarial attacks in object detection, revealing critical insights into the existing attack methods. While adversarial attacks in image classification are extensively studied, their impact on object detection presents distinct challenges that require dedicated investigation. Hence, in this paper, we reviewed different types of existing adversarial attacks on object detection and common evaluation metrics and conducted comprehensive experiments on evaluating the effectiveness and transferability of the attacks on different object detectors. Our experiments revealed several key findings: detector robustness against white-box and black-box attacks shows no correlation, and using surrogate models with matching backbones does not guarantee better transferability. Smaller objects proved more vulnerable to adversarial attacks, while newer detector architectures demonstrated improved resistance. We also found that diverse surrogate detector ensembles significantly increase attack transferability. Moving forward, key priorities include developing standardized evaluation protocols, conducting comprehensive transferability studies, and investigating the relationships between model architecture and adversarial vulnerability, particularly concerning physical-world attacks. As object detection systems increasingly power critical applications, such understanding becomes essential for building robust and reliable systems.
\section{Acknowledgement}
\label{sec: acknowledgement}

This research is supported by the Agency for Science, Technology and Research (A*STAR) under the Singapore Aerospace Programme (Grant No. M2215a0067) and the National Research Foundation of Singapore under its AI Singapore Programme (AISG Award No.: AISG2-TC-2021-003). (Corresponding authors: Khoi Nguyen Tiet Nguyen, Wenyu Zhang, and Liangli Zhen)


\bibliographystyle{ieeetr}
\bibliography{references}

\end{document}